\documentclass[10pt,journal,compsoc]{IEEEtran}
\newif\ifpeerreview

\peerreviewfalse

\usepackage[nocompress]{cite}
\usepackage{url}
\usepackage{amsmath,amssymb,graphicx}
\usepackage{multirow}
\usepackage{color}

\usepackage{bm}
\usepackage{lipsum} 

\usepackage[switch]{lineno}

\def\Sec{Section~}

\newcommand{\argmin}{\mathop{\rm arg~min}\limits}

\definecolor{modifyRebut}{rgb}{0,0,0}
\definecolor{modifyRebut2}{rgb}{0,0,0}

\newcommand{\paperID}{45}

\title{Spectral MVIR: Joint Reconstruction of\\3D Shape and Spectral Reflectance}

\author{Chunyu~Li, Yusuke~Monno, 
        and~Masatoshi~Okutomi
\IEEEcompsocitemizethanks{\IEEEcompsocthanksitem All authors are with Department of Systems and Control Engineering, School of Engineering, Tokyo Institute of Technology, Tokyo, Japan.\protect\\
E-mail:lchunyu@ok.ctrl.titech.ac.jp\\
This work was partly supported by JSPS KAKENHI Grant Number 17H00744 and 21K17762}
}

\begin{document}

\IEEEtitleabstractindextext{%
\begin{abstract}
Reconstructing an object's high-quality 3D shape with inherent spectral reflectance property, beyond typical device-dependent RGB albedos, opens the door to applications requiring a high-fidelity 3D model in terms of both geometry and photometry. In this paper, we propose a novel Multi-View Inverse Rendering~(MVIR) method called Spectral MVIR for jointly reconstructing the 3D shape and the spectral reflectance for each point of object surfaces from multi-view images captured using a standard RGB camera and low-cost lighting equipment such as an LED bulb or an LED projector. Our main contributions are twofold: (i)~We present a rendering model that considers both geometric and photometric principles in the image formation by explicitly considering camera spectral sensitivity, light's spectral power distribution, and light source positions. (ii)~Based on the derived model, we build a cost-optimization MVIR framework for the joint reconstruction of the 3D shape and the per-vertex spectral reflectance while estimating the light source positions and the shadows. Different from most existing spectral-3D acquisition methods, our method does not require expensive special equipment and cumbersome geometric calibration. Experimental results using both synthetic and real-world data demonstrate that our Spectral MVIR can acquire a high-quality 3D model with accurate spectral reflectance property.
\end{abstract}

}

\ifpeerreview
\linenumbers \linenumbersep 15pt\relax 
\author{Paper ID \paperID\IEEEcompsocitemizethanks{\IEEEcompsocthanksitem This paper is under review for ICCP 2021 and the PAMI special issue on computational photography. Do not distribute.}}
\markboth{Anonymous ICCP 2021 submission ID \paperID}%
{}
\fi
\maketitle
\thispagestyle{empty}

\IEEEraisesectionheading{
  \section{Introduction}\label{sec:introduction}
}
%
%
%
%

\IEEEPARstart{3}{D} shape and spectral reflectance are inherent geometric and photometric characteristics of an object. Typically, the photometric property of an object's 3D model reconstructed from imagery is represented as color in an RGB space. However, RGB values do not represent the inherent physical property of the object since they depend on the spectral sensitivity of a camera. In contrast, the spectral reflectance, which is defined in wavelength by wavelength, is inherent to the object and can provide much richer information about the object's surface compared with device-dependent RGB values. Therefore, acquiring a high-quality 3D model with the spectral reflectance property is valuable for many applications requiring high-fidelity information in terms of both geometry and photometry, such as digital archiving~\cite{chane2013integration,kim2014hyper3d}, artwork authentication~\cite{polak2017hyperspectral}, material classification~\cite{brusco2006system,liang2014remote}, plant modeling~\cite{liang20133d}, and relighting~\cite{wilkie2002tone}.


\begin{figure*}[t]
  \centering
 \includegraphics[width=0.9\hsize]{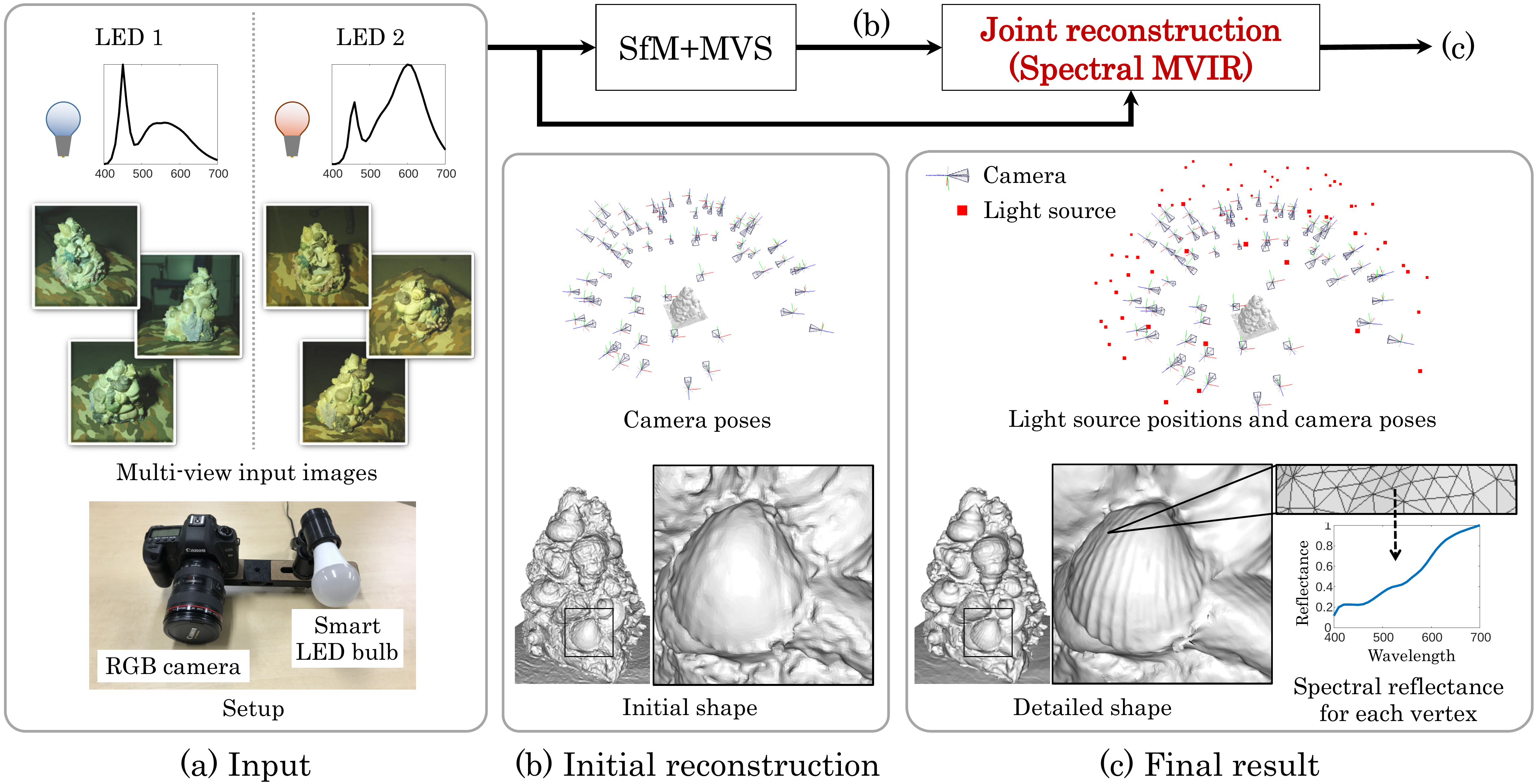}\\
 \vspace{-2mm}
  \caption{Our Spectral MVIR framework using an example camera and light setup. Using multi-view images captured with a standard RGB camera and a low-cost smart LED bulb, we first estimate camera poses and an initial mesh model based on SfM and MVS. Then, we jointly optimize each vertex's 3D position and spectral reflectance while estimating light source positions. This framework enables geometric-calibration-free and high-quality spectral-3D data acquisition.}
  \label{fig:introduction}
\end{figure*}

3D reconstruction and spectral reflectance estimation are well-studied research topics in computer vision. However, these two research fields have progressed separately. 3D reconstruction from multi-view images is becoming a basic tool owing to many useful softwares based on structure from motion~(SfM) and multi-view stereo~(MVS)~\cite{openmvs,Schoenberger,VisualSFM}. However, SfM and MVS methods only focus on geometric 3D reconstruction. Some advanced methods~\cite{maurer2016combining,wu2011high,xu2017shading}, including multi-view inverse rendering~(MVIR)~\cite{kim2016multi,Zhao}, jointly estimate a 3D shape and surface reflectance by combining the SfM and MVS pipeline with photometric 3D reconstruction using shading information. However, these methods usually focus on estimating RGB albedos, which are device-dependent and not inherent to the object. On the other hand, the spectral reflectance of an object is recovered based on multispectral imaging methods~\cite{Baek,Cao,Han2,Cui,Monno,Park,Oh}. However, these methods commonly focus on 2D multispectral imaging and ignore the geometric relationship between the object and the light source.

Some studies have proposed a system for spectral-3D data acquisition~\cite{hirai2016measuring,ito20163d,kim20123d,kitahara2015simultaneous,ozawa2017hyperspectral,zia20153d}. However, existing systems require expensive special equipment, such as a multispectral camera~\cite{kim20123d,ozawa2017hyperspectral,zia20153d} and a multispectral light source~\cite{hirai2016measuring,ito20163d,kitahara2015simultaneous}, or they need cumbersome geometric calibration for each considered setup~\cite{hirai2016measuring,kim20123d}.
In addition, most of existing systems simply add spectral information obtained by a multispectral setup to a separately estimated 3D model and do not consider the geometric information for spectral reflectance estimation, which causes the baked-in artifacts of the shading and the shadows.

A recently proposed state-of-the-art system called Pro-Cam SSfM~\cite{li2019pro} has realized geometric-calibration-free acquisition of the spectral-3D data by using an off-the-shelf RGB camera and LED projector.
This system combines multi-view structured-light and SfM techniques to simultaneously estimate the 3D points, the camera poses, and the projector poses. Then, it incorporates the geometric relationship between the 3D points and the projector positions to estimate the spectral reflectance while eliminating the effect of the shading and the shadows.
By this manner, the 3D reconstruction and the spectral reflectance estimation are performed as separated steps assuming that the active 3D scanning by structured light provides enough high-quality 3D model. However, the quality of the estimated 3D model is in fact restricted by the resolution of the projector, which is usually much lower than that of the camera.



In this paper, we propose a novel method called Spectral MVIR for jointly reconstructing a detailed 3D mesh model and the spectral reflectance for each mesh vertex.
Figure~\ref{fig:introduction} shows the overview of the Spectral MVIR framework with an example camera and light setup. To capture multi-view input images, we use a standard RGB camera and a low-cost smart LED bulb that can emit several types of illumination with different spectral power distributions\footnote{Note that, instead of a smart LED bulb, we can use separate light sources with arbitrary positions for our method.}. Using the captured input images, we first estimate camera poses and an initial mesh model based on SfM and MVS, and then jointly optimize each vertex's 3D position and spectral reflectance by minimizing multi-view and multispectral rendering errors. Our main contributions are summarized as follows. 




\begin{itemize}
    \item We present an image rendering model considering the photometric relationship among illumination's spectral power distribution, object's spectral reflectance, and camera's spectral sensitivity. We also model the geometric relationship between the object's surface and the light source position, where we adopt a near point light model, which is suitable for low-cost LED-based multispectral data acquisition. Using the derived rendering model, we can explicitly consider the effect of the shading and the shadows on the spectral reflectance estimation.
    \vspace{1mm}
    \item Based on the derived rendering model, we propose Spectral MVIR to jointly optimize each mesh vertex's 3D position and spectral reflectance while estimating the light source positions for each input image. 
    This joint optimization approach makes our system free from geometric calibration and applicable to less-constrained input images captured using an RGB camera and an LED light source.
\end{itemize}

Experimental results using both synthetic and real data demonstrate that our proposed Spectral MVIR can reconstruct a higher-quality 3D model than an existing baseline MVS method while simultaneously acquiring accurate spectral reflectance information. It is also validated that our joint reconstruction outperforms the simple combination of an MVS method and a spectral reflectance estimation method in terms of both 3D model quality and spectral reflectance accuracy. Furthermore, we apply Spectral MVIR to the initial 3D model obtained by Pro-Cam SSfM~\cite{li2019pro}, where an LED projector is used as a light source, and demonstrate that our joint reconstruction derives a refined 3D model by exploiting the shading information observed by the camera, which has a much higher pixel resolution than the projector. We also show the application of spectral-3D relighting, which reproduces the image under an arbitrary light position and spectral power distribution.

\section{Related Work}
\noindent
\textbf{Multispectral Imaging Systems:}
Many software-based~\cite{Aeschbacher,Arad,Fu,Jia4,Nguyen2,Shi} and hardware-based~\cite{Baek,Cao,Cui,Han2,Monno,Park,Oh} systems have been proposed for multispectral imaging to recover accurate spectral information. Among these systems, lighting-based systems~\cite{Cui,Han2,Park,Hidaka} have demonstrated a better trade-off between accuracy and cost since they can observe multispectral measurements using a standard RGB camera without any hardware modification by temporally changing illumination spectrum. However, existing lighting-based systems focus on 2D multispectral imaging and do not take the geometric relationship between the object surface and the light source position into account. This means that the shading and the shadow effects remain in the estimated spectral reflectance.

\vspace{2mm}
\noindent
\textbf{Spectral-3D Acquisition Systems:}
There also exist some systems for spectral-3D model acquisition. One intuitive approach is to apply an SfM and MVS pipeline using multi-view multispectral images to generate a point cloud with spectral information~\cite{ito20163d,liang20133d,miller2014passive,zia20153d}. Another approach is to combine multispectral imaging and photometric stereo to estimate dense surface normals and the spectral reflectance using the images captured under different light source positions~\cite{kitahara2015simultaneous,nam2014multispectral,ozawa2017hyperspectral}. Active 3D scanning, such as a laser scanner~\cite{kim20123d}, a ToF camera~\cite{rueda2019snapshot}, and structured light~\cite{heist20185d,hirai2016measuring}, is also combined with multispectral imaging for obtaining a high-quality 3D model with spectral information. However, the above-mentioned systems rely on expensive special equipment, such as a multispectral camera~\cite{heist20185d,kim20123d,liang20133d,miller2014passive,nam2014multispectral,ozawa2017hyperspectral,zia20153d} and a multispectral light source~\cite{hirai2016measuring,ito20163d,kitahara2015simultaneous}, or require dedicated geometric calibration for each considered system~\cite{hirai2016measuring,heist20185d,kim20123d,nam2014multispectral,rueda2019snapshot}.

A state-of-the-art system called Pro-Cam SSfM~\cite{li2019pro} has recently been proposed to realize low-cost spectral-3D data acquisition using an RGB camera and an off-the-shelf projector. Although this system is based on a careful data acquisition procedure that moves the camera and the projector alternately, it achieves geometric-calibration-free spectral-3D scanning by combining multi-view structured-light and SfM techniques. In Pro-Cam SSfM, the spectral reflectance estimation is separately performed after the 3D reconstruction assuming that the obtained 3D model with the structured light is sufficiently high quality. However, in fact, the quality of the estimated 3D model is limited by the resolution of the projector (i.e. structured light pattern), which is low compared with the resolution of the camera, resulting in the lack of detailed surface shapes. 



\vspace{2mm}
\noindent
\textbf{Advantages of Spectral MVIR:}
Compared with existing systems, our system based on Spectral MVIR has several advantages. First, it only requires a standard RGB camera and a low-cost lighting device such as an LED bulb to capture input images. Second, the data acquisition is more convenient since our system is geometric-calibration-free and allows hand-held image capturing. Third, the 3D shape and the spectral reflectance can be obtained precisely by jointly optimizing them while estimating the light positions and considering the effect of shading and shadows. Since Spectral MVIR can fully exploit a high camera resolution, more detailed shapes can be obtained when it is applied to the initial 3D model obtained with Pro-Cam SSfM.

\section{Proposed Spectral MVIR}

In this section, we detail our Spectral MVIR, which jointly estimates a detailed 3D mesh and the spectral reflectance for each mesh vertex from multi-view images. We here explain each step of Spectral MVIR assuming the setup using an LED bulb as a light source, as shown in Fig.~\ref{fig:introduction}. We will demonstrate the results using an LED projector later in the experimental result part.

\subsection{Data Acquisition}
\label{sec:dataacquisition}

As shown in Fig.~\ref{fig:introduction}, we use a standard RGB camera and a smart LED bulb to capture multi-view input images. We exploit the smart LED bulb as our light source because of two reasons: (i)~It can emit several types of illumination with different spectral power distributions and thus can be used for multispectral observations; (ii)~It is very low cost (less than twenty dollars). We fix the camera and the LED bulb on a plate and capture the images by moving the plate around an object while changing the illumination spectrum. Please note that our method does not require exactly the same viewpoint images under different illumination spectrum. Thus, hand-held image acquisition is also allowed.


\subsection{Camera Pose Estimation and Initial Reconstruction}
\label{sec:initialshape}
Camera pose estimation and dense point cloud reconstruction is firstly performed by a standard SfM and MVS pipeline~\cite{openmvs,Schoenberger,VisualSFM}.
As our method needs an initial mesh for camera visibility and cast shadow calculation, surface reconstruction~\cite{Kazhdan,Kazhdan2} is then applied to the dense point cloud to generate a mesh model.

The visibility of each mesh vertex from each camera is then checked using the estimated camera poses by SfM. If the vertex $i$ is reprocjected into the view frustum of a camera and its sight ray is not occluded by any other mesh triangle, this vertex is considered as visible from this camera. The camera which is visible from $i$-th vertex is added to the visible camera set $\mathcal{V}(i)$. Finally, the initial coarse mesh is subdivided by $\sqrt{3}$-subdivision~\cite{kobbelt20003} to obtain a sufficiently high-resolution mesh model for shape refinement.

\subsection{Joint Reconstruction of 3D Shape, Spectral Reflectance, and Light Source Positions}

\begin{figure*}[t!]
  \centering
 \includegraphics[width=0.85\hsize]{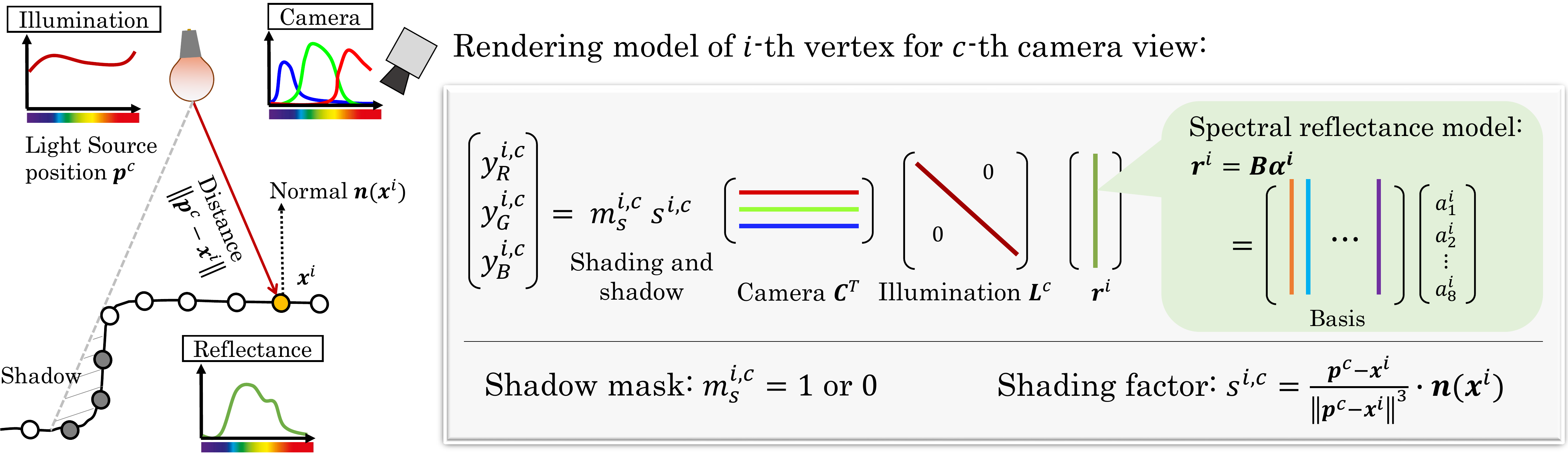}
  \caption{The illustration of our rendering model for $i$-th vertex and $c$-th camera viewpoint. The rendering value is modeled by the shadow mask, the shading factor, the camera spectral sensitivity matrix, the illumination spectral distribution matrix, and the spectral reflectance vector, where the spectral reflectance is expressed by eight basis functions. The shadow mask and the shading factor are calculated from the geometric relationship between the light source position and the 3D vertex position.}
  \label{fig:model}
\end{figure*}

The joint optimization of the 3D shape and the spectral reflectance is then performed while simultaneously estimating the light source positions. The cost function is expressed as
\begin{equation}
\begin{aligned}
\argmin_{\bm{X},\bm{A},\bm{P}}~&E_{\rm ren}(\bm{X},\bm{A},\bm{P})+ \gamma_1 E_{\rm ssm}(\bm{A})+ \gamma_2 E_{\rm psm}(\bm{X},\bm{A})\\
&+ \gamma_3 E_{\rm gsm}(\bm{X}),
\label{eq:costfunction}
\end{aligned}
\end{equation}
where $E_{\rm ren}$, $E_{\rm ssm}$, $E_{\rm psm}$, and $E_{\rm gsm}$ represent a rendering term, a spectral smoothness term, a photometric smoothness term, and a geometric smoothness term, respectively. The parameters $\gamma_1$, $\gamma_2$ and $\gamma_3$ represent the weights to balance each term's influence. The optimization parameters are described as follows:
\begin{itemize}
 \item
 $\bm{X}=\{\bm{x}^1,...,\bm{x}^{N_v}\}$ is the set of 3D vertex positions $\bm{x}^i \in \mathbb{R}^{3}$, where $N_v$ is the total number of vertexes.
 \vspace{1mm}
 \item
 $\bm{A}=\{\bm{\alpha}^1,...,\bm{\alpha}^{N_v}\}$ is the set of weight vectors for spectral reflectance basis $\bm{\alpha}^i~\in~\mathbb{R}^{N_{b}}$, where $N_{b}$ is the number of basis functions. We represent the spectral reflectance as the weighted sum of a small number of spectral basis functions, as we will explain later.
 \vspace{1mm}
 \item
 $\bm{P}=\{\bm{p}^1,...,\bm{p}^{N_c}\}$ is the set of light source positions $\bm{p}^c \in \mathbb{R}^{3}$, where $N_c$ is the total number of camera viewpoints.
\end{itemize}
In what follows, we detail each term of Eq.~(\ref{eq:costfunction}).

\subsubsection{Rendering Term}
\label{sec:renderingterm}

The rendering term $E_{\rm ren}$ evaluates the differences between observed and rendered pixel intensities. The cost function is described as
\begin{equation}
E_{\rm ren}(\bm{X}, \bm{A},\bm{P})=\sum_i^{N_v} \sum_{c \in \mathcal{V}(i)} \frac{\|\bm{y}^{i,c}_{obs}-\bm{y}(\bm{x}^i,\bm{\alpha}^i,\bm{p}^c)\|^2}{|\mathcal{V}(i)|},
 \label{eq:costren}
\end{equation}
where $\mathcal{V}(i)$ is the visible camera set for $i$-th vertex and $\bm{y}^{i,c}_{obs} \in \mathbb{R}^3$ is the observed RGB intensity vector for the pixel corresponding to the projection of $i$-th vertex to $c$-th viewpoint's image. $\bm{y} \in \mathbb{R}^3$ is the corresponding rendered intensity, which is derived based on our rendering model as detailed below.

As illustrated in Fig.~\ref{fig:model}, in our setup, each viewpoint's image is captured using one LED light source. By assuming linear camera responses, the intensity~$y$ of $n$-th camera channel for $i$-th vertex captured from $c$-th camera viewpoint is modeled as 
\begin{equation}
y^{i,c,n}=m_{s}^{i,c}s^{i,c}\int_{\Omega_\lambda}c^{n}(\lambda)l^{c}(\lambda)r^i(\lambda)d\lambda,
\label{eq:imageformulation}
\end{equation}
where $r^i(\lambda)$ is the spectral reflectance for $i$-th vertex, $l^{c}(\lambda)$ is the illumination's spectral distribution for $c$-th viewpoint, and $c^{n}(\lambda)$ is the camera spectral sensitivity for $n$-th channel ($n \in \{R,G,B\}$), $\lambda$ represents the wavelength, and $\Omega_\lambda$ is the target wavelength range, which is set to [400nm, 700nm].

Unlike conventional 2D multispectral imaging, our rendering model considers the shadow mask $m_{s}^{i,c}$ and the shading factor $s^{i,c}$, which can be calculated from the geometric relationship between the mesh vertex and the light source position. As illustrated in Fig.~\ref{fig:model}, the shadow mask $m_{s}^{i,c}$ is defined as a binary value determining whether the ray from the light source position $\bm{p}^c$ to $i$-th vertex position $\bm{x}^i$ is occluded ($m_{s}^{i,c}=0$) or not ($m_{s}^{i,c}=1$). The shading factor $s^{i,c}$ is expressed as
\begin{equation}
s^{i,c}= l_{ref}^{i,c}/l^c,
\label{eq:shading}
\end{equation}
where the shading is modeled as the irradiance ratio of the light reaching the camera $l_{ref}^{i,c}$ to the light emitted from the light source $l^c$. Note that the shading factor is wavelength independent and only determined by the geometric relationship between the light source position $\bm{p}^c$ and the 3D vertex position $\bm{x}^i$. Thus, the wavelength $\lambda$ can be omitted in the description.

If we assume Lambertian reflectance, $l_{ref}$ is independent of the camera position and expressed as
\begin{equation}
l_{ref}^{i,c}=l_{inc}^{i,c} \frac{\bm{p}^c-\bm{x}^i}{\|\bm{p}^c-\bm{x}^i\|} \cdot \bm{n}(\bm{x}^i),
\label{eq:lambartian}
\end{equation}
where $l_{inc}^{i,c}$ is the irradiance of the incident light at $i$-th vertex, $\bm{n}(\bm{x}^i) \in \mathbb{R}^3$ is the normal vector computed by $\bm{x}^i$ and its adjacent vertexes, and
$\frac{\bm{p}^c-\bm{x}^i}{\|\bm{p}^c-\bm{x}^i\|}$ indicates the normalized light direction.
\textcolor{modifyRebut}{Similar to some photometric stereo works~\cite{nam2018practical,higo2009hand},} by considering a near point light model and the inverse square illumination attenuation according to the distance between $\bm{p}^c$ and $\bm{x}^i$, the irradiance at $i$-th vertex is expressed as
\begin{equation}
l_{inc}^{i,c}=\frac{l^c}{\|\bm{p}^c-\bm{x}^i\|^2}.
\label{eq:inversesquare}
\end{equation}
By assuming that ambient light and interreflection are negligible, the shading factor is finally derived by substituting Eq.~(\ref{eq:lambartian}) and Eq.~(\ref{eq:inversesquare}) into Eq.~(\ref{eq:shading}) as
\begin{equation}
s^{i,c}=\frac{\bm{p}^c-\bm{x}^i}{\|\bm{p}^c-\bm{x}^i\|^3}\cdot \bm{n}(\bm{x}^i).
 \label{eq:finalshading}
\end{equation}

\begin{figure*}[t!]
  \centering
 \includegraphics[width=0.9\hsize]{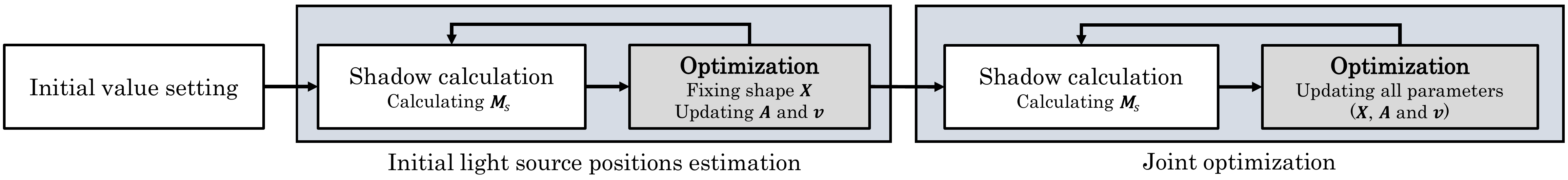}\\ \vspace{-3mm}
   \caption{Optimization steps. The initial light source positions calculated from~$\bm{v}$ are firstly estimated by fixing the 3D vertex positions~$\bm{X}$. Then,  all the parameters (3D vertex positions~$\bm{X}$, spectral reflectances~$\bm{A}$ and light source positions~$\bm{v}$) are jointly optimized using the initial light source positions. The shadow mask calculation, where $\bm{M}_s\in \mathbb{R}^{N_{v} \times N_{c}}$ is the shadow mask set for all vertexes and light source positions, and the optimization are separately performed and iterated several times.}
  \label{fig:optimizationdetails}
\vspace{-4mm}
\end{figure*}

In practice, spectral functions in Eq.~(\ref{eq:imageformulation}) are discretized with some wavelength intervals into $N_{\lambda}$ elements ($N_{\lambda}=31$ for the case of [400nm, 700nm] with 10nm intervals). Thus, Eq.~(\ref{eq:imageformulation}) can be represented in a matrix form by
\begin{equation}
\bm{y}^{i,c}= m_{s}^{i,c} s^{i,c} \bm{C}^T \bm{L}^c \bm{r}^i,
\label{eq:imageformulationmatrix}
\end{equation}
where $\bm{y}^{i,c} \in \mathbb{R}^{3}$ is the rendered RGB intensity vector, $\bm{C}^T \in \mathbb{R}^{3 \times N_{\lambda}}$ is the camera sensitivity matrix, $\bm{L}^c \in \mathbb{R}^{N_{\lambda} \times N_{\lambda}}$ is the diagonal illumination spectral distribution matrix, and $\bm{r}^i \in \mathbb{R}^{N_{\lambda}}$ is the spectral reflectance vector.

It is known that the spectral reflectance of natural objects can be represented by the weighted sum of a small number of orthogonal basis functions~\cite{Han2,Park,Oh}. By using basis functions, the spectral reflectance vector $\bm{r}^i$ in Eq.~(\ref{eq:imageformulationmatrix}) can be modeled as
\begin{equation}
\bm{r}^i=\bm{B} \bm{\alpha}^i,
 \label{eq:reflectancebasis}
\end{equation}
where $\bm{B} \in \mathbb{R}^{N_{\lambda} \times N_{b}}$ is the basis matrix, $\bm{\alpha}^i \in \mathbb{R}^{N_{b}}$ is the weight vector, and
$N_{b}$ is the number of basis functions. In our experiment, we use 8 basis functions, which is calculated using the spectral reflectance data of 1269 Munsell color chips~\cite{Munsell} by principal component analysis.
\textcolor{modifyRebut}{We use 8 basis functions based on the observation that they are able to represent more than 99\% of the total variance of \textcolor{modifyRebut2}{common reflectance data}~\cite{Oh}.} Using spectral reflectance basis model,
recovering the spectral reflectance of a vertex is equivalent to estimating the weight vector which has less unknown parameters.
Finally, given the parameters of $\bm{x}^i$, $\bm{\alpha}^i$, and $\bm{p}^c$, 
the rendered pixel intensity is expressed as
\begin{equation}
\bm{y}^{i,c}(\bm{x}^i,\bm{\alpha}^i,\bm{p}^c)=  m_{s}^{i,c} s^{i,c} \bm{C}^T \bm{L}^c \bm{B} \bm{\alpha}^i.
\label{eq:imageformulationbasis}
\end{equation}

Using the rendering term, our \textcolor{modifyRebut2}{global optimization loss takes into account both} photometric information (i.e., shading) and 3D information (i.e., disparity) to refine the 3D shape.
Specifically, the shading is calculated \textcolor{modifyRebut2}{using the} surface normal and light source distance which are derived from estimated 3D vertex position $\bm{x}^i$ and light source position $\bm{p}^c$ (Eq.~(\ref{eq:finalshading})). Thus, the 3D surface can be refined by minimizing the rendering errors caused by incorrect shading.
In addition, the rendering term uses disparity to evaluate the consistency among input multi-view images.
During our optimization process, each estimated 3D vertex position is reprojected to the 2D image planes for all visible cameras. Then, the rendering error is evaluated for all reprojected vertices, so that each 3D vertex position can be refined by minimizing the rendering errors caused by incorrect reprojections.
In this manner, our method considers both shading and disparity, which are complementary to each other.


\subsubsection{Spectral Smoothness Term}

It is known that the spectral reflectance of a natural object is generally smooth. Thus, we impose a smoothness constraint on the estimated spectral reflectance. The cost of the spectral smoothness term $E_{\rm ssm}$ is defined as
\begin{equation}
E_{\rm ssm}(\bm{A})= \sum_i^{N_v} \|\bm{D}\bm{B}\bm{\alpha}^i\|^2,
 \label{eq:costssm}
\end{equation}
where $\bm{D} \in \mathbb{R}^{N_\lambda \times N_\lambda}$ is the operation matrix to calculate the second-order derivative~\cite{Park} along the wavelength direction.

\subsubsection{Photometric Smoothness Term}

There are ambiguities in separating the reflectance from the shading since our rendering model allows spatially varying spectral reflectance. The photometric smoothness term $E_{\rm psm}$ is applied to regularize this ambiguity~\cite{barron2013intrinsic,kim2016multi} as 
\begin{equation}
 E_{\rm psm}(\bm{X},\bm{A}) = \sum_{i}^{N_v}\sum_{j \in \mathcal{A}(i)} w^{i,j}({\bm x}^i,{\bm x}^j)\|\bm{B}\bm{\alpha}^i-\bm{B}\bm{\alpha}^j\|^2,
 \label{eq:costpsm}
\end{equation}
where $\mathcal{A}(i)$ is the set of adjacent vertexes of $i$-th vertex and $w^{i,j}$ is a weight to determine whether the change of the reflectance between adjacent vertexes is allowed or not. We use the same weight as~\cite{kim2016multi}, which is based on the intensity or chromaticity difference observed in the input image.

\subsubsection{Geometric Smoothness Term}
We apply the same geometric smoothness term as MVIR~\cite{kim2016multi}, which is defined as 
\begin{equation}
 E_{\rm gsm}(\bm{X}) = \sum_{i}^{N_v} \left(\frac{dist(\bm{x}^i, \Omega^i)}{l^i}\right)^2.
 \label{eq:costgsm}
\end{equation}
This term evaluates the distance $dist(\bm{x}^i, \Omega^i)$ between $i$-th vertex position $\bm{x}_i$ and a local plane $\Omega^i$ computed from its adjacent vertexes. $l^i$ is a normalizing parameter to discount the scene scale (see~\cite{kim2016multi} for details).

\subsection{Optimization Steps}
\label{sec:optimizationdetails}

\begin{figure*}[t]
    \centering
    \begin{minipage}[b]{0.42\hsize}\vspace{0pt}
      \includegraphics[width=0.40\hsize]{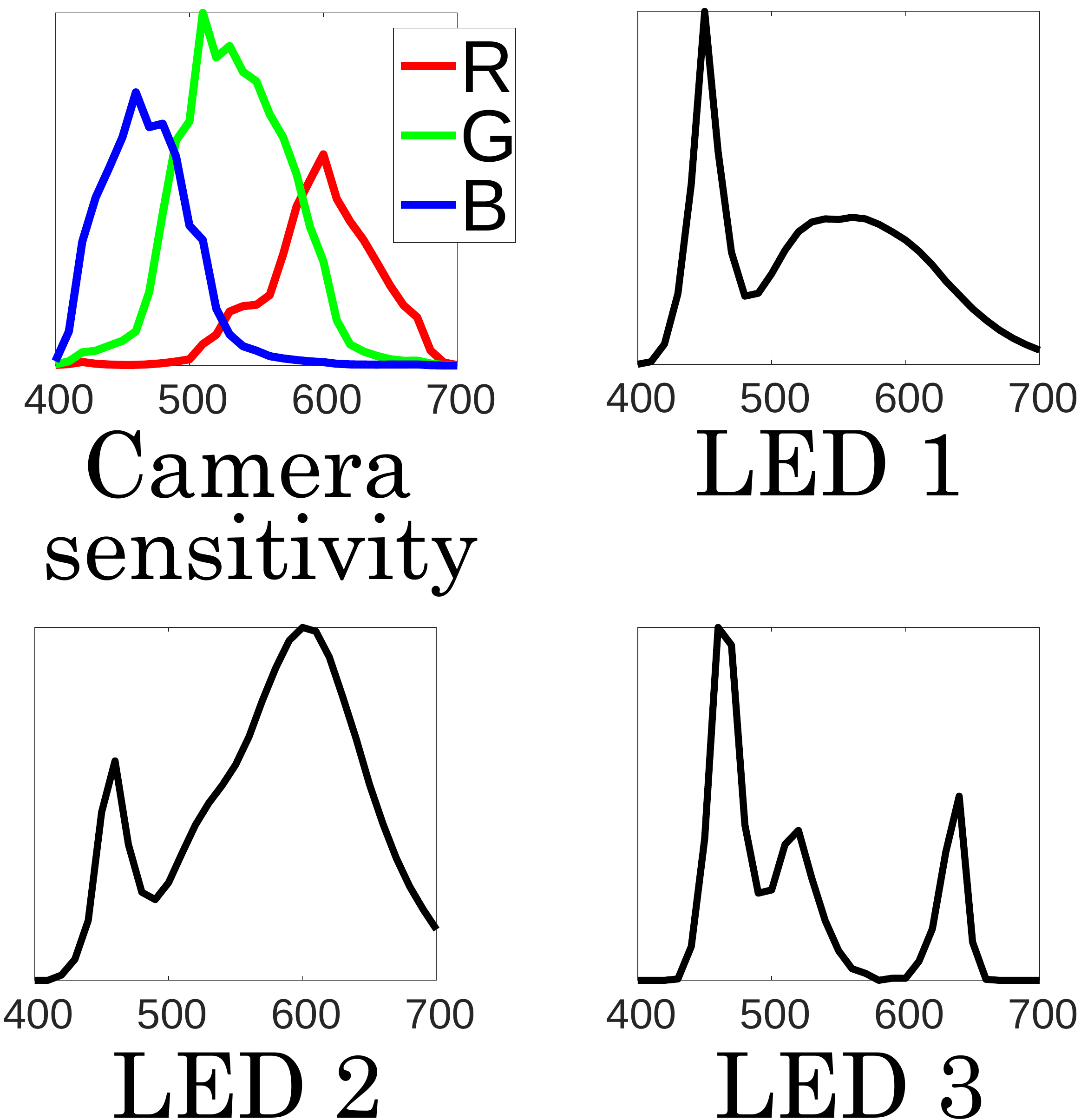}
      \includegraphics[width=0.58\hsize]{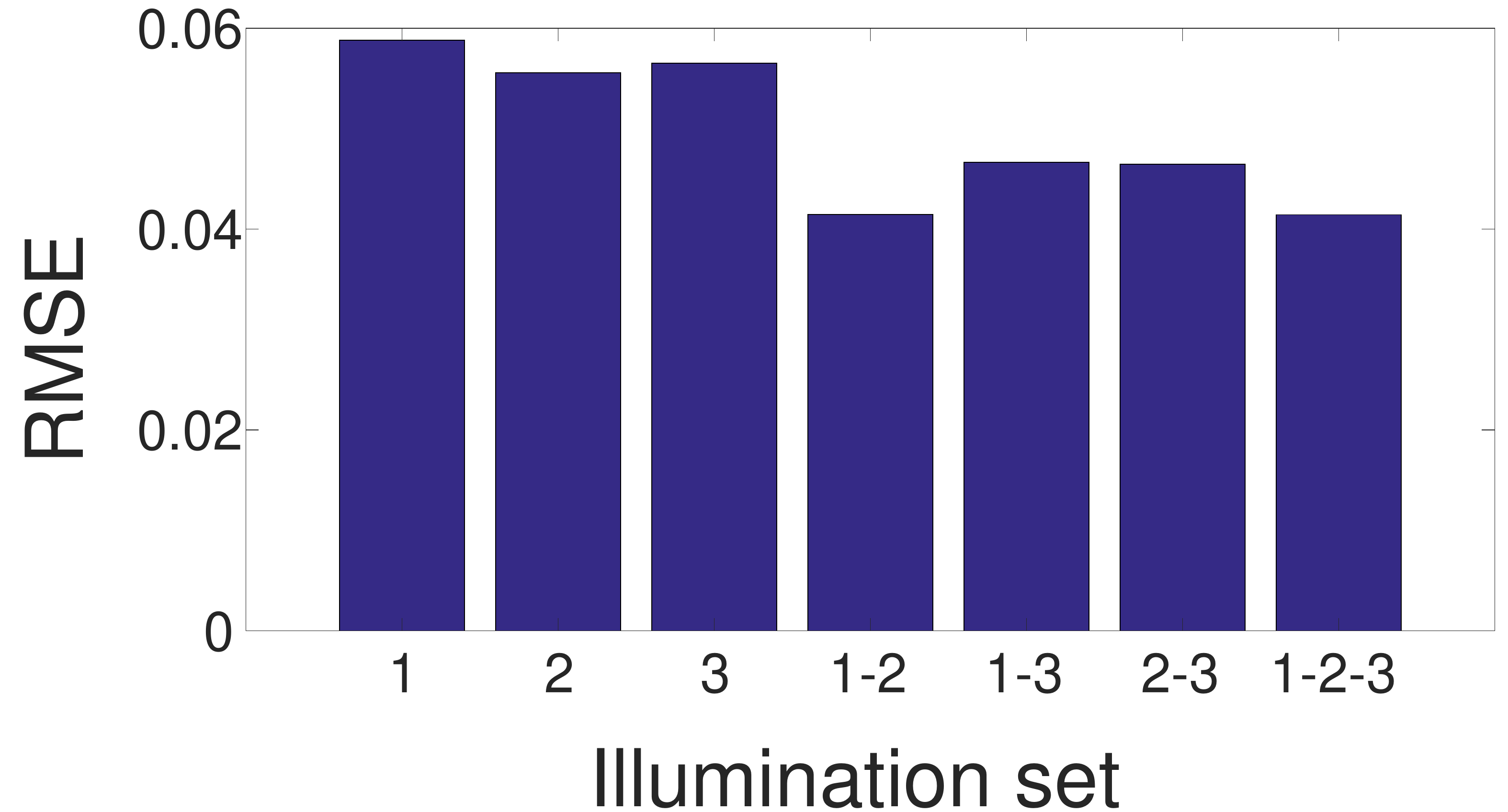}
      \caption{Left:~Camera spectral sensitivity and LED illumination's spectral power distributions. Right: Average spectral reflectance RMSE for the 24 patches of the colorchart when using different illumination combinations.}\label{fig:illuminationselection}
    \end{minipage}
    \hspace{1mm}
    \begin{minipage}[b]{0.56\hsize}\vspace{0pt}
    \centering
    \begin{minipage}{0.49\hsize}
    \centering
    \includegraphics[width=\hsize]{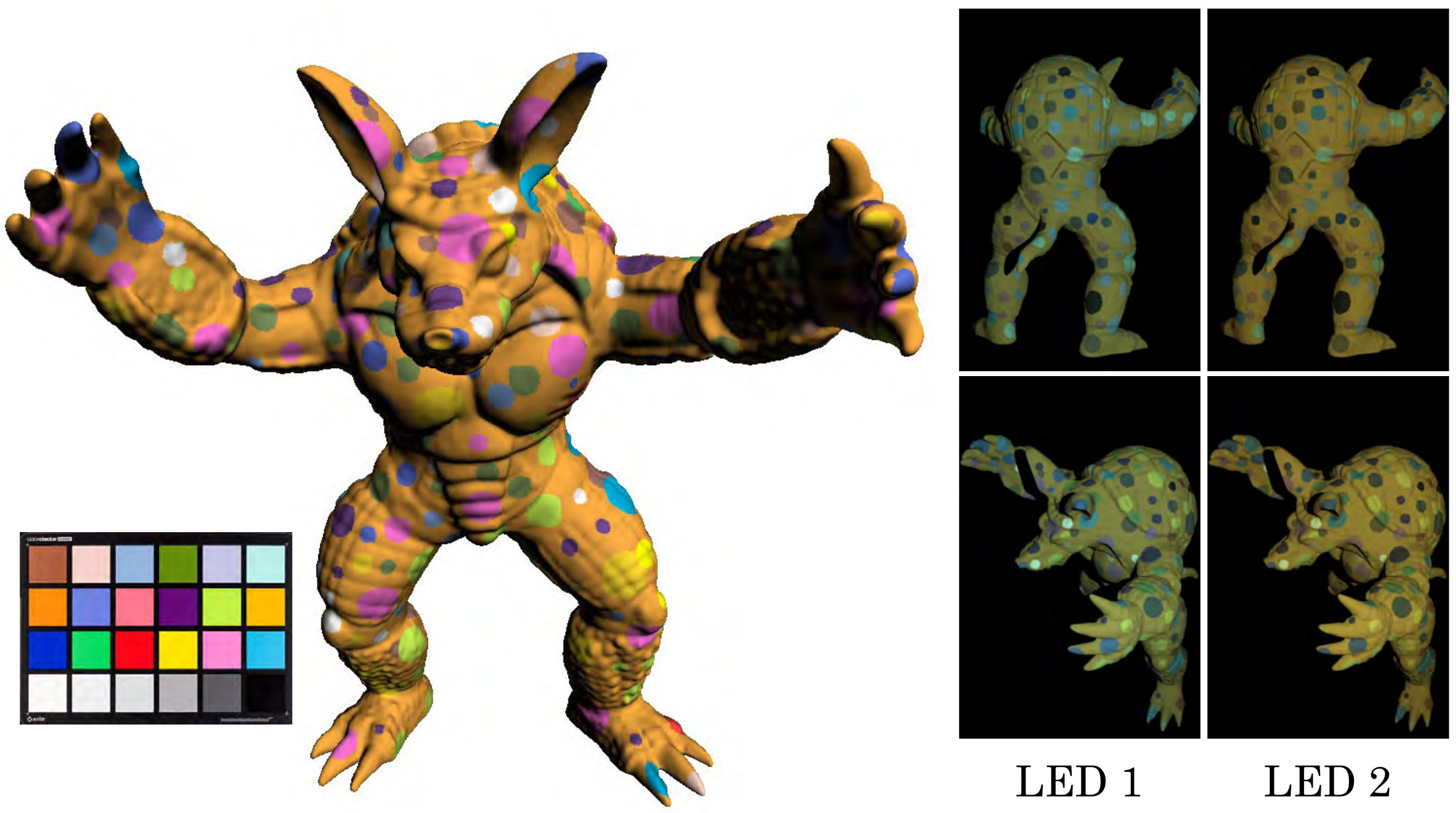}\\
     \footnotesize{(a) Armadillo} 
    \end{minipage}
    \begin{minipage}{0.49\hsize}
    \centering
    \includegraphics[width=\hsize]{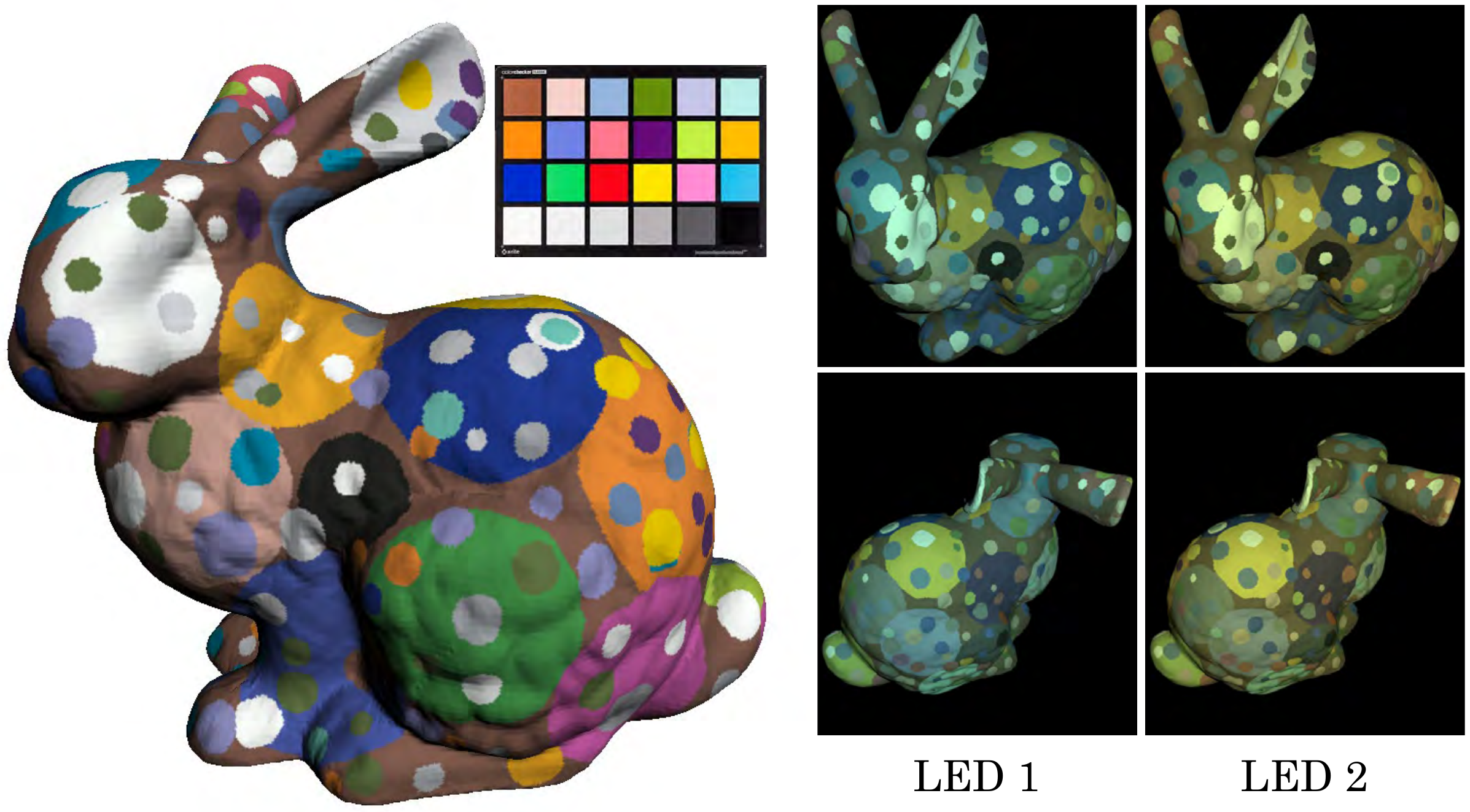}\\
    \footnotesize{(b) Stanford Bunny}
    \end{minipage}
      \caption{Synthetic spectral-3D models of (a)~Armadillo and (b)~Stanford Bunny. Left:~Visualization of the 3D model with ground-truth spectral reflectance data for each 3D point. Right:~Example rendered images under two illuminations.}\label{fig:mvssetup}
    \end{minipage}
\end{figure*}

\begin{figure*}[ht]
  \centering
    \begin{minipage}[t!]{0.26\hsize}
 \includegraphics[width=\hsize]{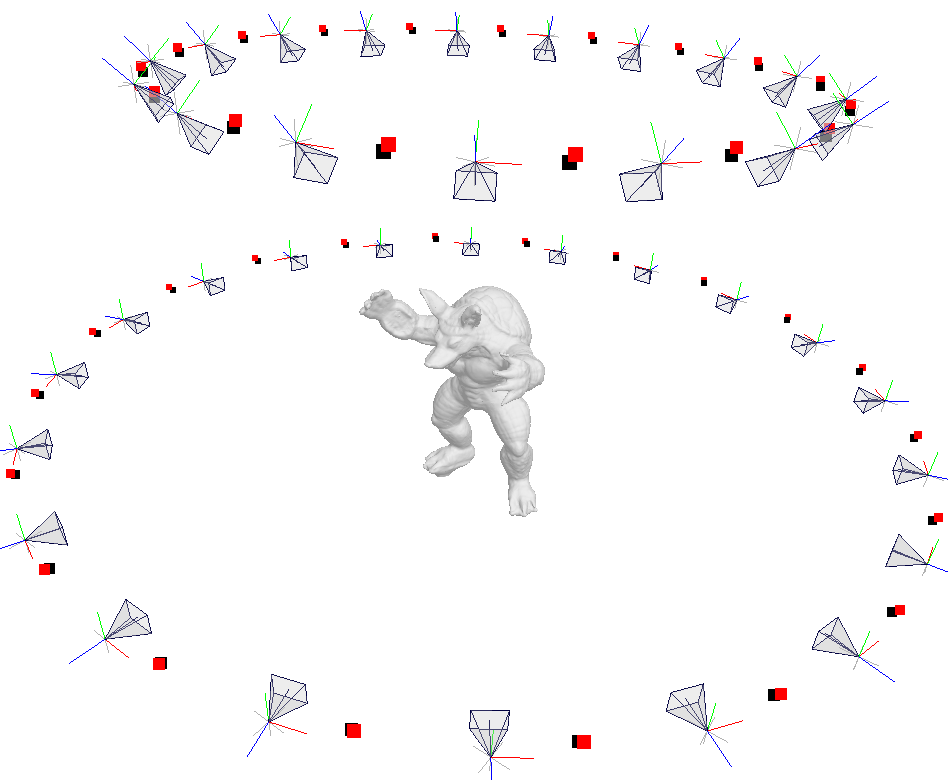}
  \caption{Ground-truth camera poses (pyramids), ground-truth light source positions (black points), and estimated light source positions by our method (red points) for Armadillo.}
  \label{fig:syntheticlight}
    \end{minipage}
    \hspace{1mm}
    \begin{minipage}[t!]{0.72\hsize}\vspace{0pt}
 \includegraphics[width=\hsize]{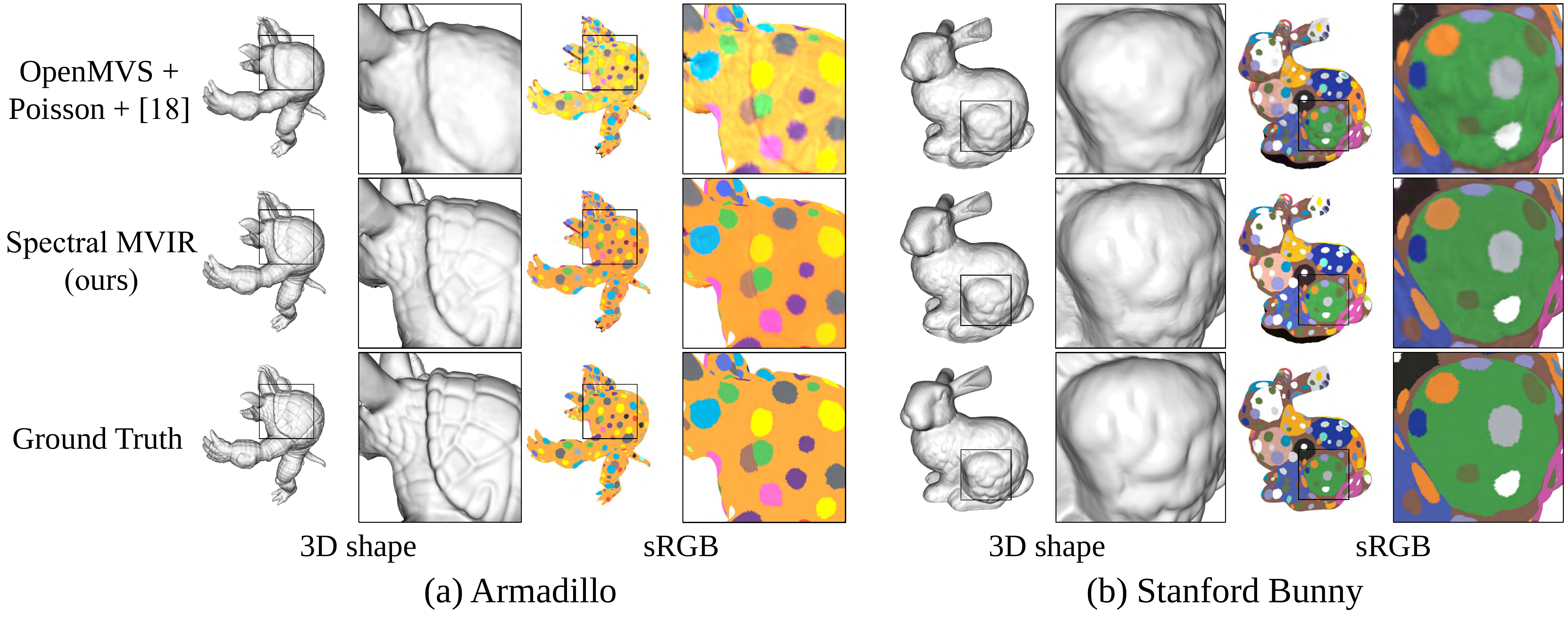}\\ \vspace{-8mm}
  \caption{Qualitative comparison of the estimated 3D shapes and the sRGB color representations converted from the estimated spectral reflectances.}
  \label{fig:mvsqualitative}
    \end{minipage}
\end{figure*}

In our setup described in \Sec\ref{sec:dataacquisition}, the camera and the LED light source are rigidly fixed on a plate. Thus, it is only necessary to estimate the translation vector $\bm{v} \in \mathbb{R}^3$, which is defined in the camera coordinate as the vector from the origin to the light source position. Using the translation vector, the light source position $\bm{p}^c$ corresponding to $c$-th camera viewpoint can be calculated as
\begin{equation}
 \bm{p}^c=({\bm{R}^c})^{-1}(\bm{v}-\bm{t}^c),
 \label{eq:optimizationdetails}
\end{equation}
where $\bm{R}^c$ and $\bm{t}^c$ are extrinsic parameters of camera viewpoint $c$ which can be derived from SfM. Since the translation vector is the same for all viewpoints, we only need to estimate the translation vector for estimating  all light positions.

Figure~\ref{fig:optimizationdetails} shows our optimization steps. Firstly, the initial vector for $\bm{v}$ is set as~${\bm 0}$ (i.e. the origin of the camera coordinate). To avoid a local minimum solution, we then estimate a proper initial light source positions by fixing the 3D vertex positions during the optimization. Finally, we optimize all parameters using the initial light source positions. Since the calculation of the shadow mask $m_{s}$ for one vertex requires the whole estimated mesh, the optimization and the shadow calculation are performed separately and iterated several times until the parameters converge.

The weights in Eq.~(\ref{eq:costfunction}) are empirically set as $\gamma_1=0.01$, $\gamma_2=2$, and $\gamma_3=0.01$.
The non-linear optimization problem of Eq.~(\ref{eq:costfunction}) is solved using C++ Ceres Solver~\cite{ceressolver}.

\section{Experimental Results}

We conducted the experiments using two setups. As the light source, the first setup uses an LED bulb as shown in Fig.~\ref{fig:introduction}, while the second setup uses an LED projector as in Pro-Cam SSfM~\cite{li2019pro}. \textcolor{modifyRebut}{For both setups, we assume that the camera sensitivity and the illumination spectrum \textcolor{modifyRebut2}{have been calibrated.}} The experimental details are explained below.

\subsection{Camera and LED Bulb Setup}
\label{sec:comparisonMVS}

\subsubsection{Setup Details}
\label{sec:mvssetup}

In this setup, we used a Haodeng WH018 RGBCW smart LED bulb as the light source. This LED bulb can emit three types of illumination with different spectral power distributions, \textcolor{modifyRebut}{which were measured by using a StellarNet BlueWave-VIS Spectrometer, as shown in Fig.~\ref{fig:illuminationselection}.}
For capturing input images, we used a Canon EOS 5D Mark-II digital camera.
The camera spectral sensitivity of this model was obtained from the database of~\cite{Jiang}, as shown in Fig.~\ref{fig:illuminationselection}.

Before conducting the spectral-3D reconstruction experiments, we selected the best illumination set for spectral reflectance estimation using the spectral reflectance data of the 24 patches of an X-Rite's colorchart. Figure~\ref{fig:illuminationselection} shows the average RMSE comparison for the estimated spectral reflectance of the 24 patches when using different illumination combinations. It can be observed that RMSE is reduced by using two or three illuminations
, validating that multispectral observations improve the spectral reflectance estimation accuracy. In our experiments, we used the illumination set of (LED 1, LED 2) since there is no big difference between the result using this set and the result using all three illuminations.
\textcolor{modifyRebut2}{For our optimization problem, two illuminations result in information from only six spectral bands (i.e., 2 illuminations $\times$ 3 color channels), fewer than the dimensionality of the spectral basis functions ($N_b = 8$ in our experiments). To ensure that the inverse problem we solve is well-posed, we regularize our optimization through the spectral smoothness term $E_{\rm ssm}$ and the photometric smoothness term $E_{\rm psm}$.}

\begin{figure*}[ht]
  \centering
    \begin{minipage}[t!]{0.49\hsize}\vspace{0pt}
 \includegraphics[width=\hsize]{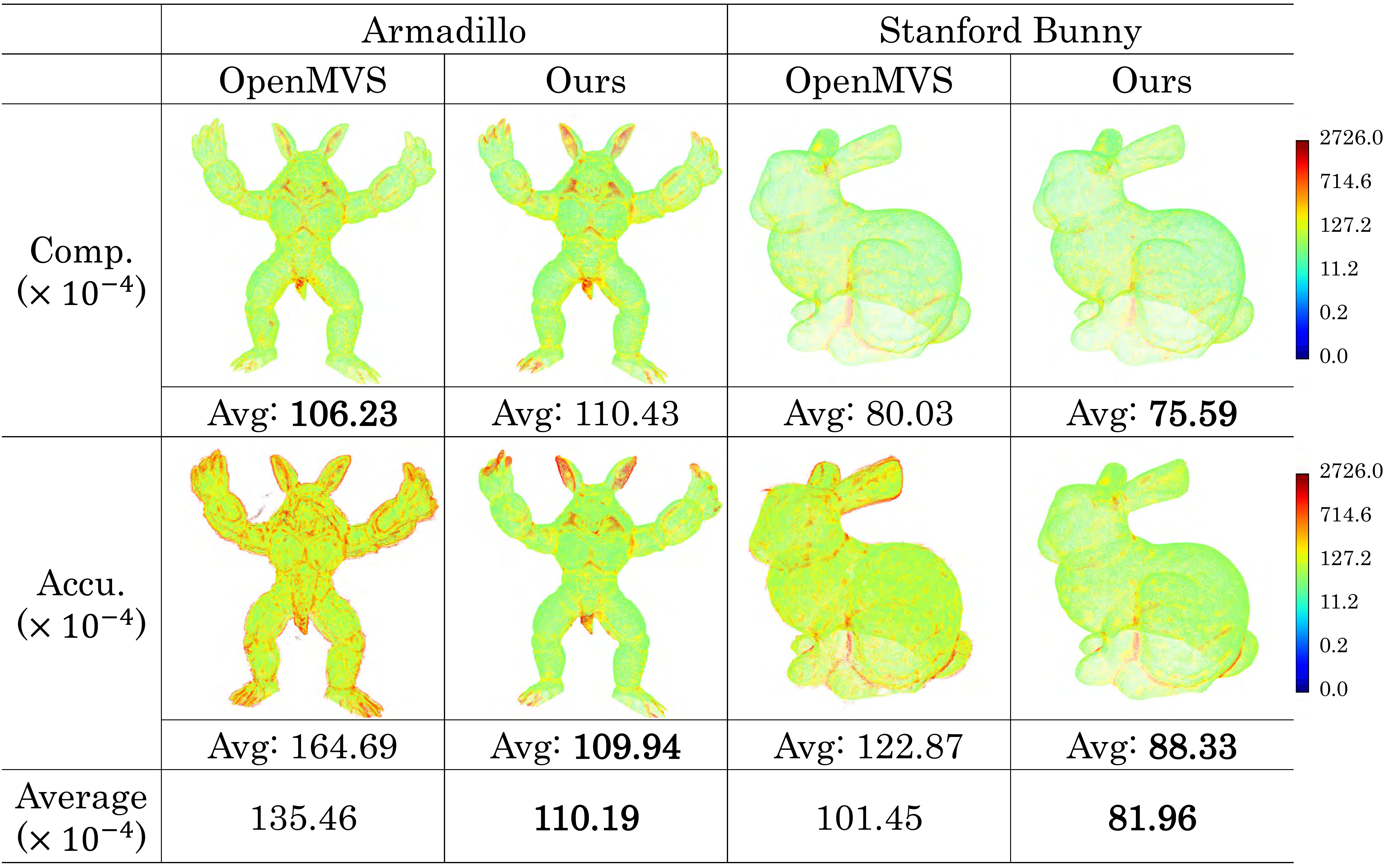}\\ \vspace{-2mm}
  \caption{Quantitative comparison of the estimated 3D shapes using the error metrics of the completeness and the accuracy~\cite{ley2016syb3r} (lower is better).}
  \label{fig:mvsquantitative}
    \end{minipage}
    \hspace{1mm}
    \begin{minipage}[t!]{0.49\hsize}\vspace{0pt}
 \includegraphics[width=\hsize]{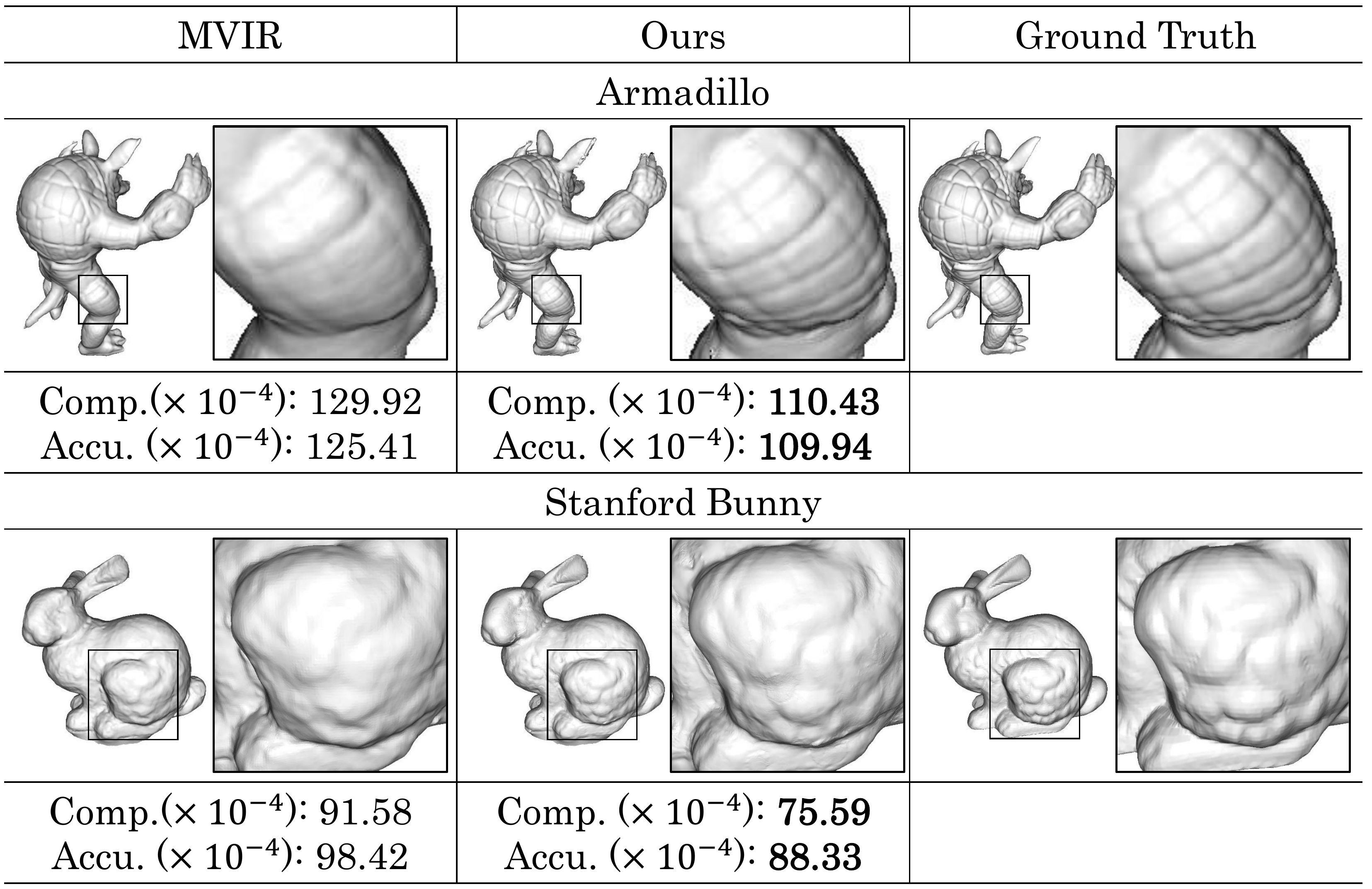}\\ \vspace{-2mm}
  \caption{\textcolor{modifyRebut}{Comparison of the estimated 3D shapes with MVIR~\cite{kim2016multi}.}}
  \label{fig:mvir}
    \end{minipage}
\end{figure*}

\begin{figure*}[t!]
  \centering
  \begin{minipage}[b]{0.33\hsize}\vspace{0pt}
    \centering
    \includegraphics[width=\hsize]{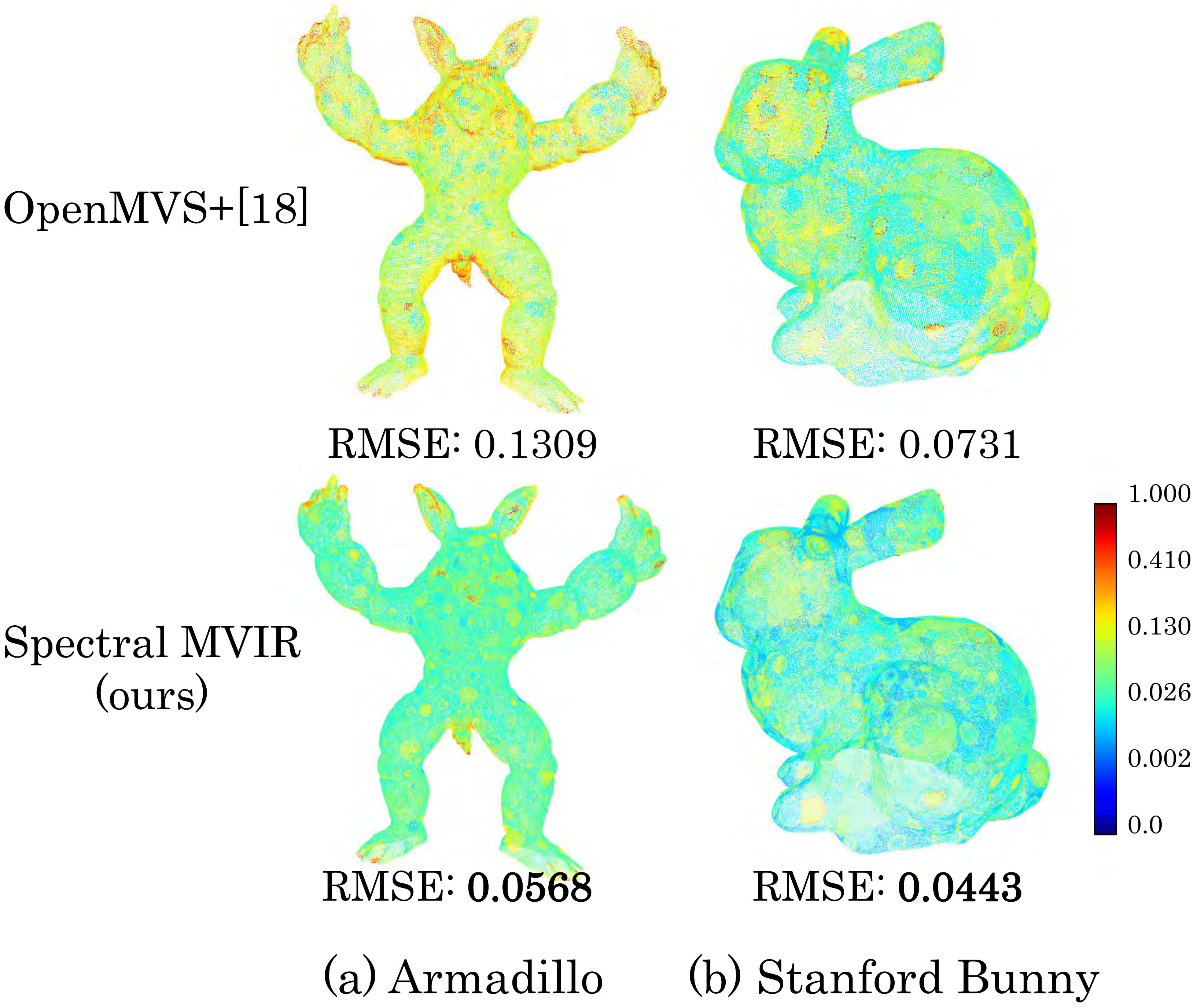}
  \caption{Spectral reflectance error (RMSE over all wavelengths) for each 3D point on (a) Armadillo and (b) Stanford Bunny.}
  \label{fig:mvsquantitativereflectance}
  \end{minipage}
  \hspace{1mm}
  \begin{minipage}[b]{0.63\hsize}\vspace{0pt}
    \centering
    \includegraphics[width=\hsize]{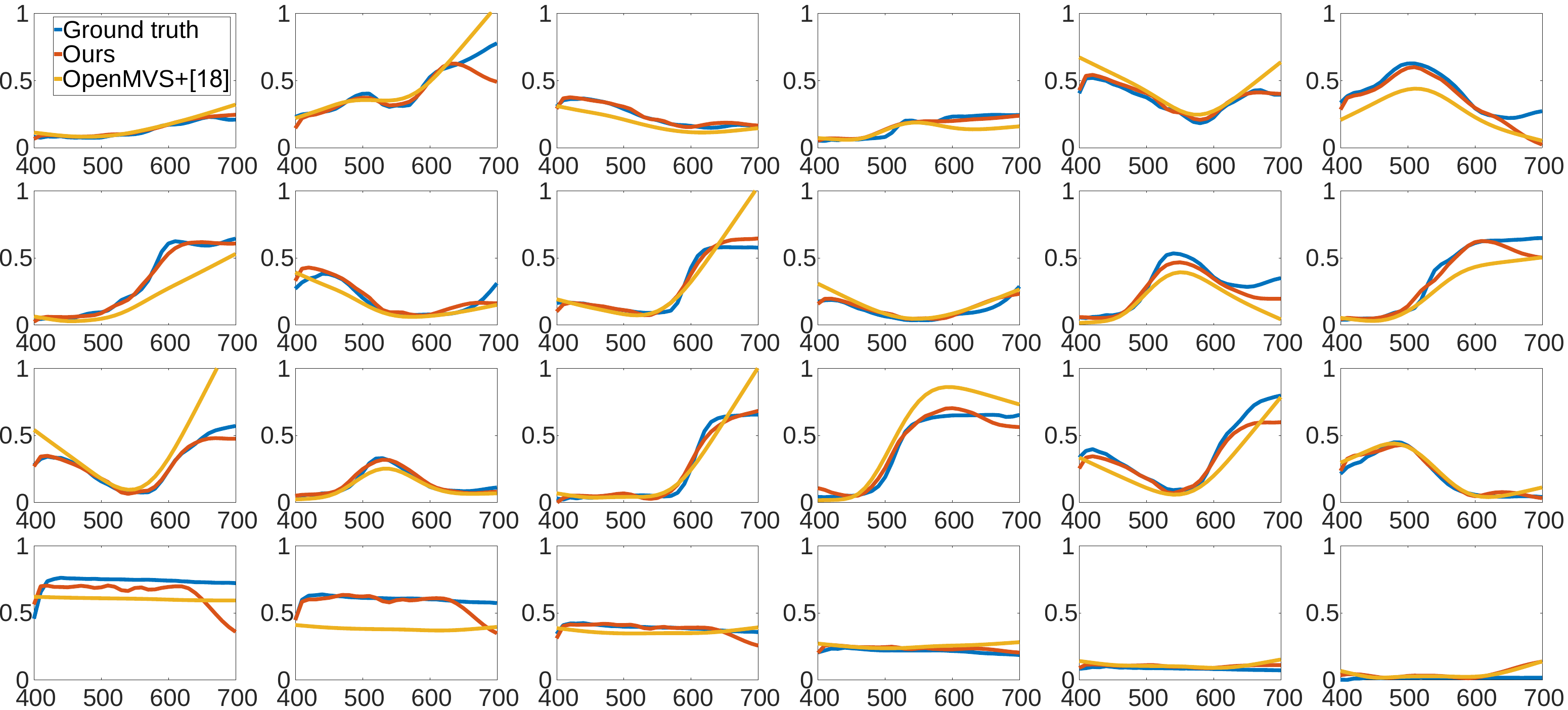}
  \caption{Sampled spectral reflectance results for each of colorchart's 24 spectral reflectance data on Armadillo. The blue line is the ground truth, the red line is our result (average within each patch), and the yellow line is the results of the combined method~(OpenMVS+\cite{Han2}).}
  \label{fig:syntheticreflectance}
  \end{minipage}
\end{figure*}

\subsubsection{Simulation Results}
\label{sec:mvssimulation}

To quantitatively evaluate the 3D shape quality and the spectral reflectance accuracy, we preformed simulation experiments using two CG models (Armadillo and Stanford Bunny) obtained from Stanford 3D Scanning Repository~\cite{stanford}. Using each CG model, we created a ground-truth spectral-3D model, which has ground-truth spectral reflectance data for every 3D point. As the ground-truth spectral reflectance data, we used the spectral reflectance data of the colorchart's 24 patches. According to the texture shown in Fig.~\ref{fig:mvssetup}, each reflectance data of the 24 patches was assigned to the 3D points. Using the created ground-truth spectral-3D model, multi-view input images with the resolution of 1024 $\times$ 768 were generated using 74 camera-light pair positions, as shown in Fig.~\ref{fig:syntheticlight}. For each camera-light pair position, two images were rendered using the camera and the two illuminations as described in~\Sec\ref{sec:mvssetup}. Some examples of the rendered images are shown in Fig.~\ref{fig:mvssetup}.

In the simulation experiments, we applied OpenMVS~\cite{openmvs} for initial point cloud reconstruction, where we used the ground-truth camera poses, and then applied Poisson surface reconstruction~\cite{Kazhdan2} to generate the initial mesh model for our Spectral MVIR. We compared our method with the combination of OpenMVS and an existing image-based spectral reflectance estimation method~\cite{Han2}, since there is no existing spectral-3D reconstruction method directly applicable to the considered camera and LED bulb setup, to the best of our knowledge. For the spectral reflectanc estimation of the compared method, we first applied the image-based spectral reflectance estimation method~\cite{Han2} to every input image. Then, we calculated the spectral reflectance of each vertex of the generated OpenMVS's mesh by averaging the spectral reflectance results estimated at all the pixels corresponding to the 3D vertex point projection to all visible camera images.

Figure~\ref{fig:syntheticlight} shows the ground-truth light source positions (black points) and the estimated light source positions by our method (red points). It can be confirmed that our method can accurately estimate the light source positions.
Figure~\ref{fig:mvsqualitative} shows the qualitative comparison of the estimated 3D shapes and the sRGB color representations converted from the estimated spectral reflectances.
We can confirm that our method successfully refines the initial model reconstructed by OpenMVS by exploiting shading ques.
Also, our sRGB results are closer to the ground truth and represent the object's inherent property less affected by the shading and the shadows compared with the sRGB results of the combined method~(OpenMVS+\cite{Han2}).

\textcolor{modifyRebut2}{As discussed in \Sec{\ref{sec:renderingterm}}, spectral MVIR considers both disparity and shading information for 3D shape refinement. In Fig.~\ref{fig:mvir}, we compare our method with MVIR~\cite{kim2016multi}, which is one of the state-of-the-art 3D shape refinement methods also considering both disparity and shading. We can confirm that our method recovers finer details than MVIR; this is because our method estimates the point light source positions and shadows, whereas MVIR estimates a general spherical harmonics illumination model.}


\begin{figure*}[t!]
  \centering
  \vspace{2mm}
 \includegraphics[width=0.9\hsize]{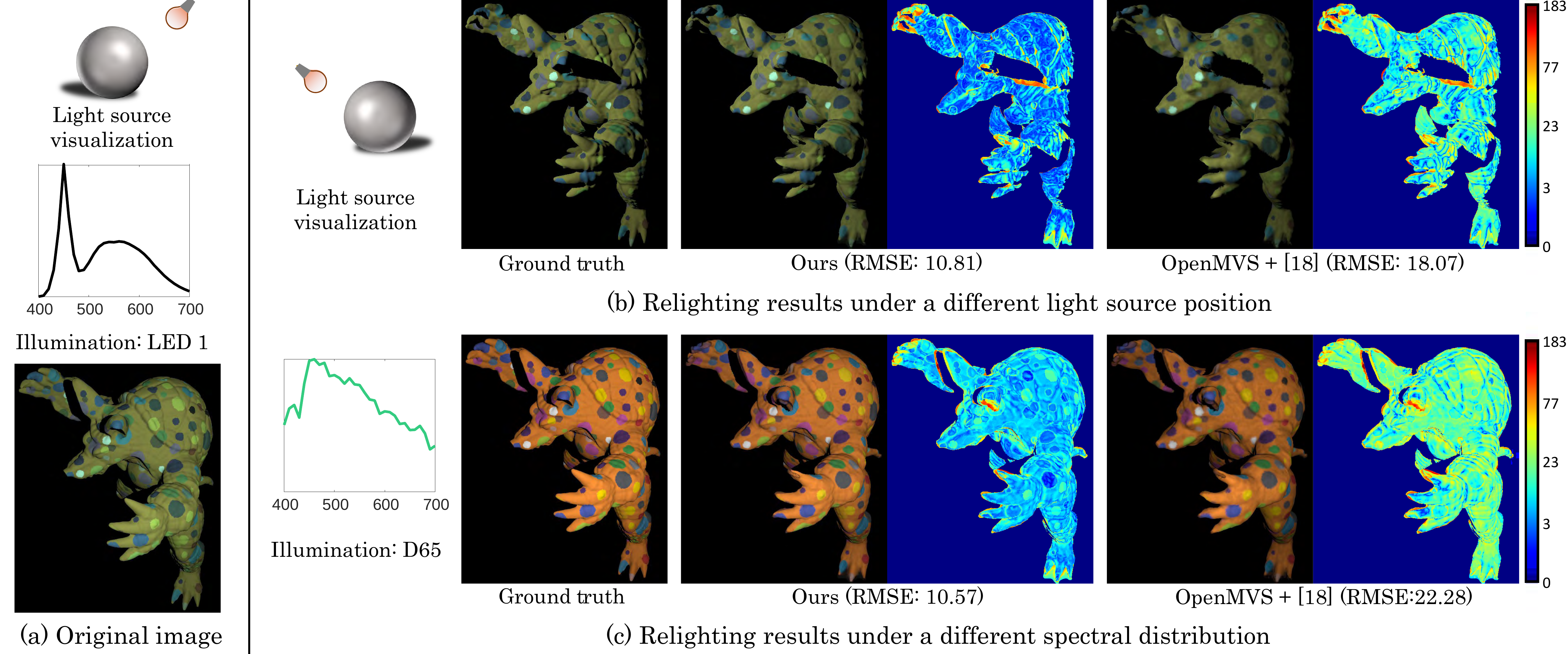}\\ \vspace{-2mm}
  \caption{Spectral-3D relighting results for Armadillo. (a)~An original image, which is one of the captured input images; (b) Relighting results under a different light source position; (c) Relighting results under a different spectral distribution.}
  \label{fig:syntheticrelighting}
\end{figure*}

\begin{figure*}[t!]
\centering
 \includegraphics[width=0.9\hsize]{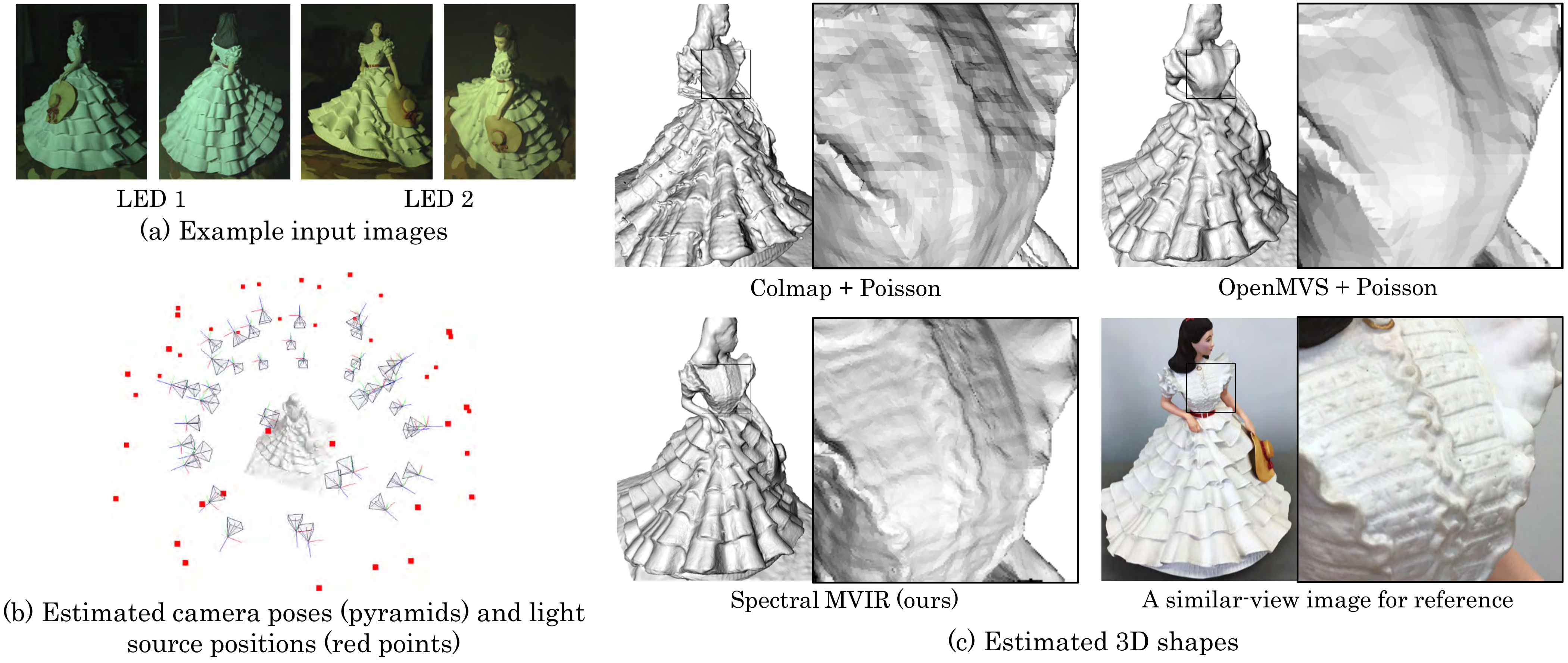}\\
\vspace{-3mm}
  \caption{3D reconstruction results for a clay sculpture. (a) Example input images; (b) Estimated camera poses and light source positions by our method; (c) Comparison of the estimated 3D shapes.}
  \label{fig:shapescarlet}
\vspace{-4mm}
\end{figure*}

To evaluate the 3D shape quantitatively, we used two common metrics, i.e. completeness and accuracy~\cite{aanaes2016large,ley2016syb3r}. The completeness is the distance from each ground-truth 3D point to its nearest estimated 3D point. The accuracy is the distance from each estimated 3D point to its nearest ground-truth 3D point. 
Figure~\ref{fig:mvsquantitative} shows the comparison of the completeness and the accuracy errors, where the errors for each 3D point are colorized and the average errors are shown below the error maps.
We can see that our method achieves better accuracy on the both models. Although OpenMVS provides better completeness on the Armadillo model, it shows much worse accuracy than our method. For the average results of accuracy and completeness, our method can provide better results compared with OpenMVS.

We next evaluate the spectral reflectance accuracy. Figure~\ref{fig:mvsquantitativereflectance} shows the error map for estimated spectral reflectance, where RMSE over all wavelengths is visualized for each 3D point. We can confirm that our method achieves lower average RMSEs compared with the combined method~(OpenMVS+\cite{Han2}). This is because that the combined method does not consider the effects of shading and shadows on the spectral reflectance estimation, resulting in large errors for complicated surfaces.
Figure~\ref{fig:syntheticreflectance} shows the sampled spectral reflectance results for each of colorchart’s 24 spectral reflectance data on the Armadillo model. The blue line is the ground truth, the red line is our result, and the yellow line is the result of the combined method~(OpenMVS+\cite{Han2}). We can confirm that our method achieves closer spectral reflectances to the ground truths than the combined method~(OpenMVS+\cite{Han2}).


We next demonstrate the results of spectral-3D relighting, which is the application of reproducing the image under the light source with an arbitrary position and spectral distribution. Figure~\ref{fig:syntheticrelighting}(a) shows an original image rendered as one of the captured input images. Figure~\ref{fig:syntheticrelighting}(b) shows the relighting results under a different light source position using the estimated spectral-3D model, while Fig.~\ref{fig:syntheticrelighting}(c) shows the relighting results under a different spectral distribution. As shown in the error maps compared with the ground truth, we can confirm that our method achieves better spectral-3D relighting accuracy compared with the combined method~(OpenMVS+\cite{Han2}).  

\begin{figure*}[t!]
\centering
\begin{minipage}{0.90\hsize}
  \vspace{0mm}
    \begin{minipage}[b]{0.23\hsize}
    \centering
    \includegraphics[width=\hsize]{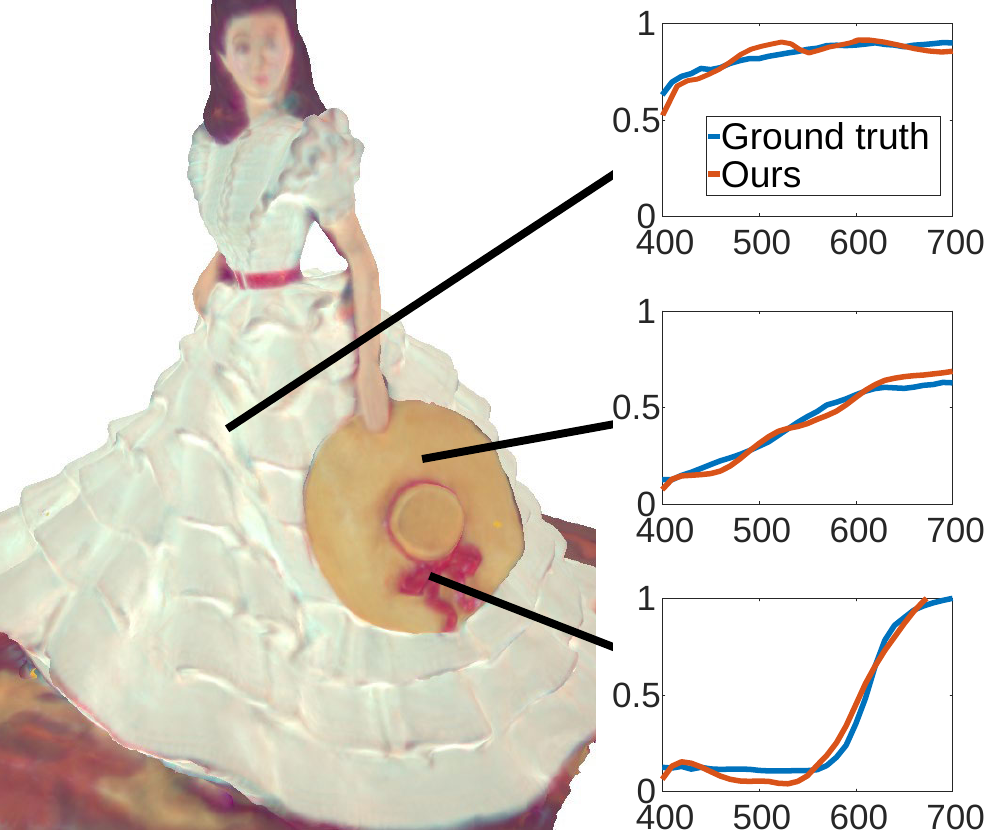}\\ \vspace{1mm}
    \scriptsize{(a) sRGB visualization and spectral reflectance results}
    \end{minipage}
    \hspace{2mm}
    \begin{minipage}[b]{0.75\hsize}
    \centering
    \includegraphics[width=\hsize]{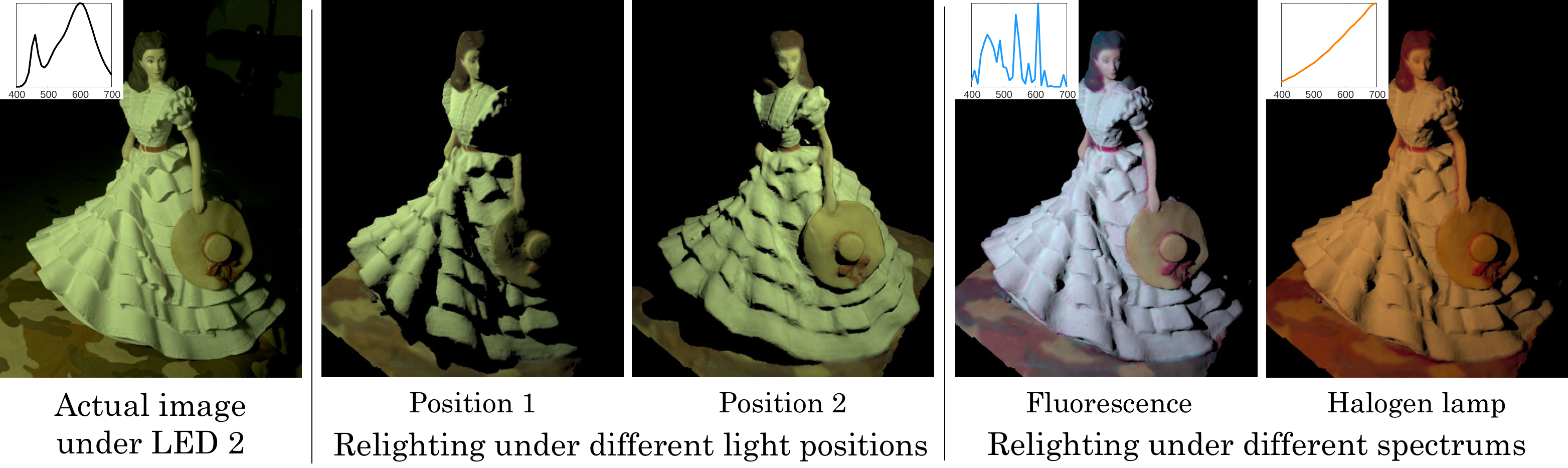}
    \scriptsize{(b) Spectral-3D relighting results}
    \end{minipage}
\end{minipage}
\caption{Spectral reflectance and relighting results for a clay sculpture. (a) The sRGB visualization and estimated spectral reflectances by our method; (b) Spectral-3D relighting results under an arbitrary light position and spectral power distribution.}
\label{fig:scarletreflectance}
\end{figure*}

\begin{figure*}[t!]
    \centering
    \includegraphics[width=0.9\hsize]{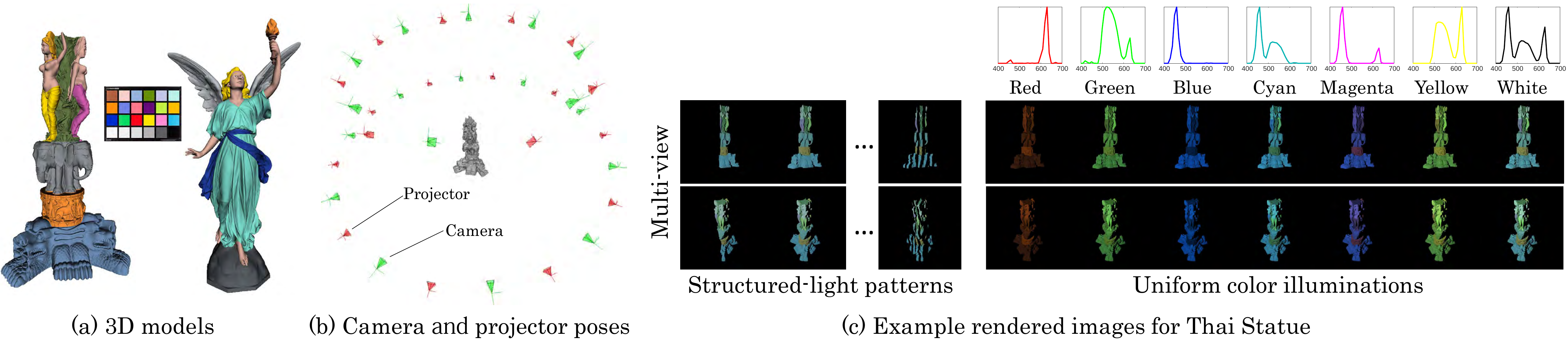}\\ \vspace{-2mm}
\caption{(a) Synthetic Thai Statue (left) and Lucy (right) models with ground-truth spectral reflectance data for each 3D point; (b) Ground-truth camera (green) and projector (red) poses for rendering input images; (c) Example rendered images under structured-light (gray-code) patterns and uniform color illuminations for Thai Statue.}\label{fig:syntheticobjects}
\vspace{-4mm}
\end{figure*}

\subsubsection{Results for Real Objects}

We next evaluate our method for real images. We captured the images of a clay sculpture from 40 viewpoints using our image acquisition setup, as shown in Fig.~\ref{fig:introduction}. Figure~\ref{fig:shapescarlet}(a) shows example input images, where each viewpoint's image was captured in the RAW format by a hand-held manner under one of two illuminations.
In our method, we applied Colmap~\cite{Schoenberger} for camera pose estimation (SfM), OpenMVS~\cite{openmvs} for dense point cloud reconstruction (MVS), and Poisson surface reconstruction~\cite{Kazhdan2} for initial mesh generation.
Figure~\ref{fig:shapescarlet}(b) shows the estimated camera poses by Colmap and the estimated light source positions by our method. 
Figure~\ref{fig:shapescarlet}(c) compares our 3D shape result with the results of Colmap's MVS and OpenMVS.
We can confirm that our Spectral MVIR can reconstruct finer details than Colmap and OpenMVS.

Figure~\ref{fig:scarletreflectance}(a) shows the sRGB result and the spectral reflectance results for some sampled vertexes estimated by our method, which are very close to ground truths measured by a \textcolor{modifyRebut}{StellarNet BlueWave-VIS} Spectrometer.
Figure~\ref{fig:scarletreflectance}(b) shows the results of spectral-3D relighting under different light source positions (middle) or different spectral power distributions (right). Those results demonstrate that we can effectively perform the spectral-3D relighting using the estimated detailed 3D shape and the spectral reflectances for every vertex.

\subsection{Camera and Projector Setup}
\label{sec:ssfmresult}

In this section, we compare our proposed method with state-of-the-art Pro-Cam SSfM~\cite{li2019pro} by using the 3D model obtained by Pro-Cam SSfM as the initial model for our joint reconstruction by Spectral MVIR.

\subsubsection{Setup Details}
\label{sec:ssfmsetup}

We applied the same setup as Pro-Cam SSfM~\cite{li2019pro}, where an ASUS P3B LED projector and a Canon EOS 5D Mark-II digital camera are used. According to the data acquisition procedure of Pro-Cam SSfM, the projector was used to project the sequence of structured-light~(gray-code) patterns and seven uniform color illuminations, as shown in Fig.~\ref{fig:syntheticobjects}(c), with the projector resolution of 1024$\times$768. \textcolor{modifyRebut}{The illumination spectral power distributions were measured by using a StellarNet BlueWave-VIS Spectrometer.} The images with the sequentially projected illuminations were captured using the EOS 5D Mark-II camera with the pixel resolution of 5616$\times$3170. The spectral sensitivity of the camera is shown in Fig.~\ref{fig:illuminationselection}. The input images for Pro-Cam SSfM were captured from multiple viewpoints by alternately moving the camera and the projector around the object with arbitrary positions, as shown in Fig.~\ref{fig:syntheticobjects}(b).

Pro-Cam SSfM sequentially performs the 3D reconstruction and the spectral reflectance estimation. In the 3D reconstruction step, the camera poses, the projector poses, and the 3D points are simultaneously estimated based on multi-view feature correspondences using the structured-light patterns. We used the 3D model estimated by Pro-Cam SSfM as the initial model for our Spectral MVIR refinement while jointly estimating the spectral reflectance. For the joint optimization, we fixed the camera poses and the projector poses to the ones estimated by Pro-Cam SSfM, since they are sufficiently accurate because of reliable feature matching using the structured-light patterns in Pro-Cam SSfM.

\subsubsection{Simulation Results}
To evaluate the 3D model refinement by Spectral MVIR, we employed two CG models (Thai Statue and Lucy) that have detailed surfaces, which were downloaded from Stanford 3D Scanning Repository~\cite{stanford}. 
As shown in Fig.~\ref{fig:syntheticobjects}(a), each ground-truth spectral-3D model with ground-truth spectral reflectance data was generated in the same way as described in \Sec\ref{sec:mvssimulation}.
Multi-view input images were generated using 18 camera and projector poses, as shown in Fig.~\ref{fig:syntheticobjects}(b). For each camera viewpoint, two sequences of images under structured-light (gray-code) patterns and uniform color illuminations were rendered by strictly following the Pro-Cam SSfM's data acquisition procedure~\cite{li2019pro}. The examples of the rendered images for two viewpoints are shown in Fig.~\ref{fig:syntheticobjects}(c).


\begin{figure*}[t!]
  \centering
 \includegraphics[width=0.9\hsize]{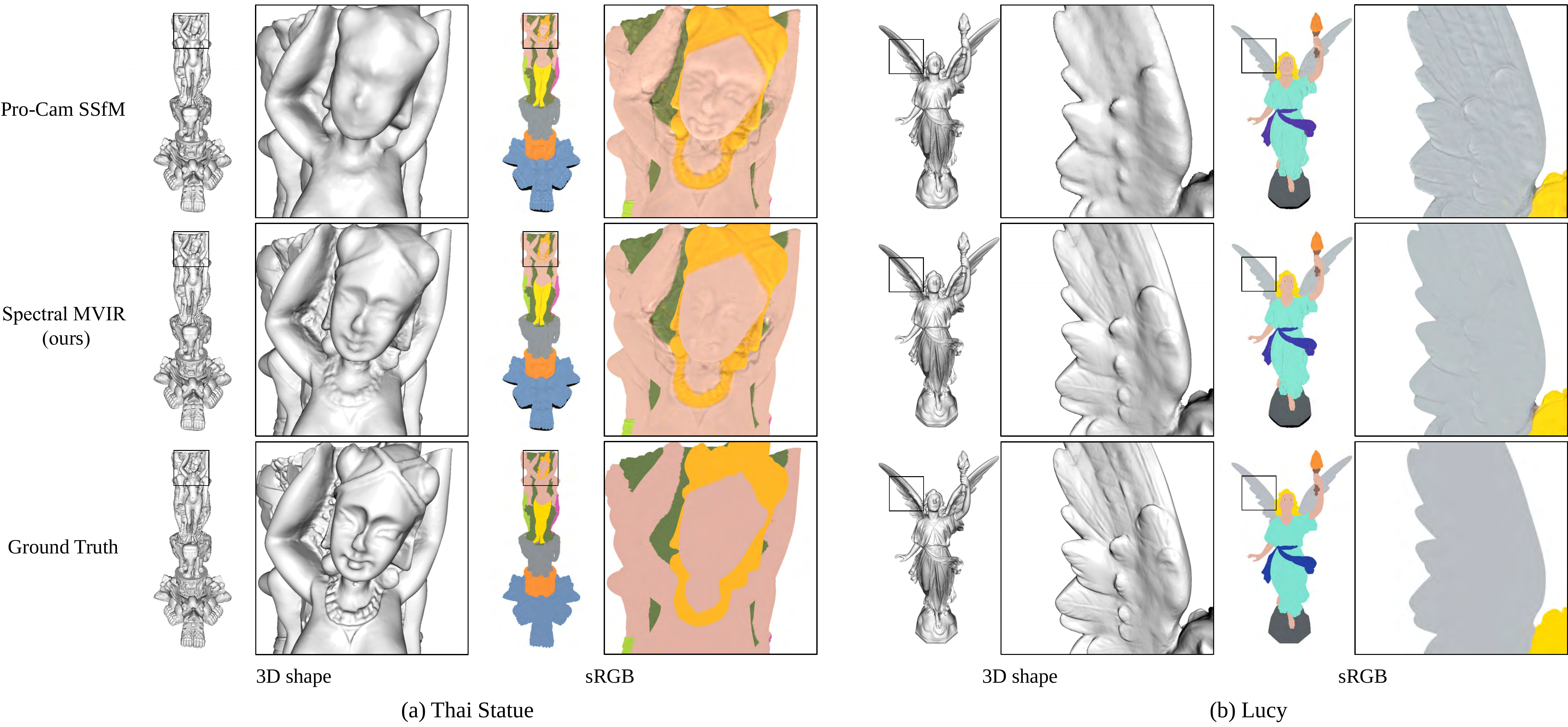}\\
  \vspace{-3mm}
  \caption{The qualitative comparison of the estimated 3D shapes and the sRGB color representations converted from the estimated spectral reflectaces. Pro-Cam SSfM~\cite{li2019pro} failed to reconstruct the detailed surfaces, which also affects the spectral reflectance accuracy as can be seen as severe ``baked-in" shading artifacts in the sRGB results.  In contrast, our Spectral MVIR can more accurately reconstruct the 3D shape and the spectral reflectance by jointly optimizing and refining them.}
  \label{fig:procamqualitative}
 \vspace{-5mm}
\end{figure*}

\begin{figure}[t!]
  \centering
 \includegraphics[width=\hsize]{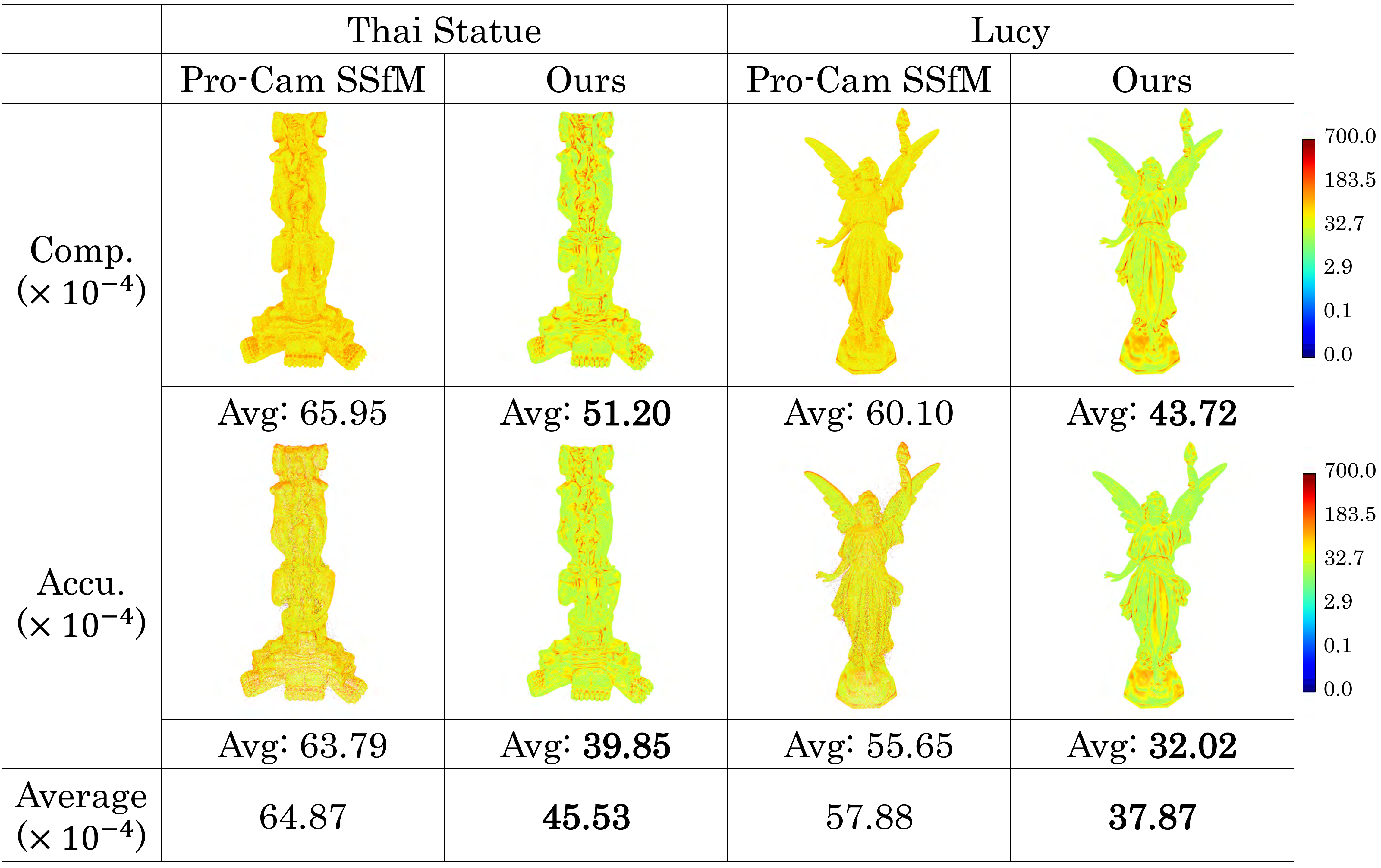}\\ \vspace{-2mm}
  \caption{The quantitative comparison of the estimated 3D shapes. The completeness and the accuracy errors are visualized in color.}
  \label{fig:procamquantitative}
 \vspace{-2mm}
\end{figure}

\begin{figure}[t!]
  \centering
  \begin{minipage}[b]{\hsize}\vspace{0pt}
  \centering
    \includegraphics[width=\hsize]{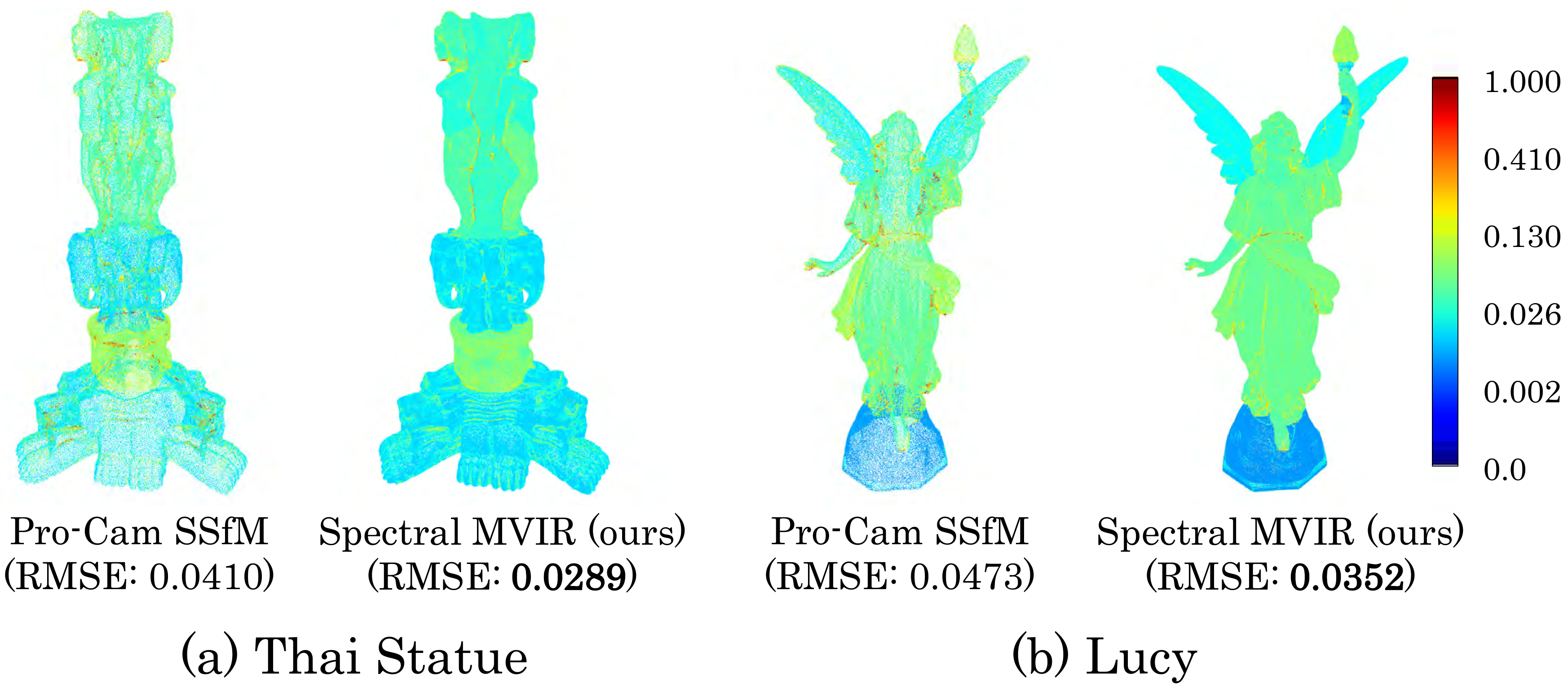}\\ \vspace{-3mm}
  \caption{The qualitative comparison of the estimated spectral reflectances. For each point, the RMSE over all wavelengths is visualized in color.}
  \label{fig:procamquantitativereflectance}
 \vspace{-2mm}
  \end{minipage}\\
\end{figure}

Figure~\ref{fig:procamqualitative} shows the qualitative comparison of the estimated 3D shapes and the sRGB color representations converted from the estimated spectral reflectances. We can see that the reconstructed 3D shape by Pro-Cam SSfM shows lower quality than the reconstructed 3D shape by our method, especially on detailed surfaces such as the face of Thai Statue and the wings of Lucy. This is the limitation of Pro-Cam SSfM, where the 3D model resolution is restricted by the projector resolution, which is low in general.
The coarse 3D reconstruction on detailed surfaces causes the error of the estimated spectral reflectance, which can be seen as severe ``baked-in" shading artifacts in the sRGB results of Pro-Cam SSfM, as shown in Fig.~\ref{fig:procamqualitative}.
In contrast, our Spectral MVIR can effectively refine the 3D model and reconstruct the surface details by using the shading information observed for high-resolution camera pixels.

Figure~\ref{fig:procamquantitative} shows the quantitative comparison of the 3D shape results using the completeness and the accuracy error metrics~\cite{aanaes2016large,ley2016syb3r}.
We can confirm that our method outperforms Pro-Cam SSfM in both the completeness and the accuracy with clear margins.
Figure~\ref{fig:procamquantitativereflectance} visualizes the RMSEs of the spectral reflectance results for each 3D point.
We can confirm that our method shows lower average spectral reflectance RMSEs, as the 3D shape refinement via the joint shape and reflectance optimization also contributes to the improvment of the spectral reflectance accuracy.


\subsubsection{Results for Real Objects}

\begin{figure}[t!]
\centering
 \includegraphics[width=\hsize]{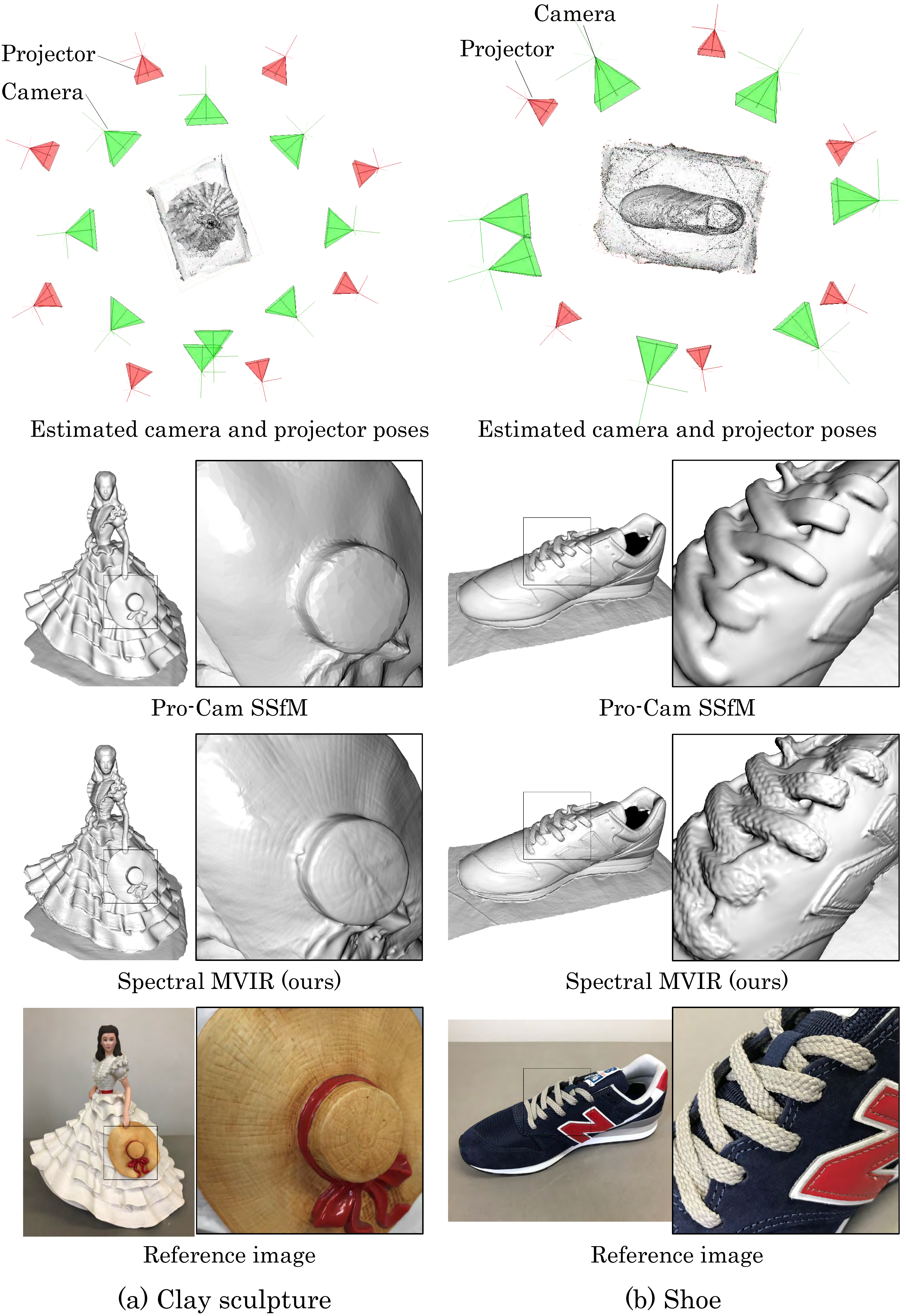}\\
 \vspace{-3mm}
  \caption{The 3D reconstruction results for two real objects, (a) a clay sculpture and (b) a shoe. From top to bottom, we show the estimated camera (green) and projector (red) poses by Pro-Cam SSfM, the estimated 3D shapes by Pro-Cam SSfM, the estimated 3D shapes by our Spectral MVIR, and color images seen from similar view points for reference.}
  \label{fig:realdataprocam}
\end{figure}

Figure~\ref{fig:realdataprocam} shows the 3D reconstruction results for two real objects, a clay sculpture (left) and a shoe (right). The top images show the results of camera and projector poses estimated by Pro-Cam SSfM, which were used to refine the 3D shape by Spectral MVIR. From the 3D shape results, we can confirm that our method can reconstruct more detailed surfaces than Pro-Cam SSfM, which is because the resolution of the reconstructed shape by Pro-Cam SSfM depends on the resolution of the projector, whereas that of the reconstructed shape by our method depends on the resolution of the camera which is much higher in general.

\subsection{Limitations}

\begin{figure}[t!]
\centering
 \includegraphics[width=\hsize]{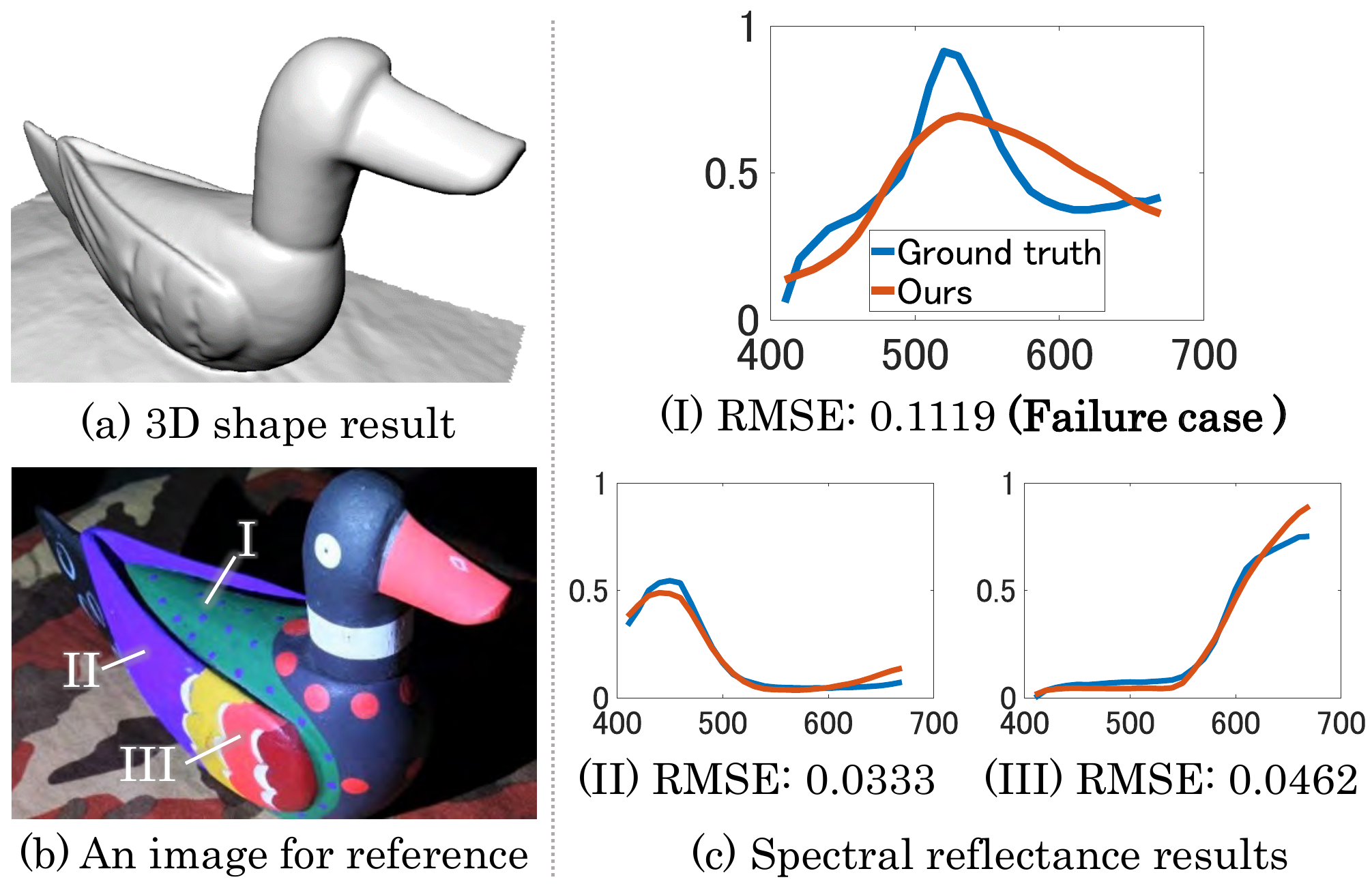}\\
 \vspace{-3mm}
   \caption{\textcolor{modifyRebut}{A failure case on an object with a narrow spectral reflectance curve. (a) The estimated 3D shape using a projector as the light source. (b) An image for reference. (c) The estimated spectral reflectance results. The narrow spectral reflectance curve (I) that cannot be represented by smooth basis functions is not estimated accurately, compared with more smooth spectral reflectance curves (II and III).}}
  \label{fig:failurecase}
\end{figure}

Spectral MVIR still has several limitations. First, we assume that the camera sensitivity and the illumination spectrum are known or pre-estimated, which requires the spectral calibration of the system. Second, the spectral basis model brings smooth reflectance results for narrow spectral reflectance curves that cannot be represented by smooth basis functions. Figure~\ref{fig:failurecase} shows a failure case, where the narrow spectral reflectance curve (I) is not recovered accurately, compared with more smooth spectral reflectance curves (II and III). Third, Spectral MVIR is currently limited to the objects with Lambertian reflectance and cannot handle objects with more complex reflectances, such as glossy objects including specular reflections\textcolor{modifyRebut2}{, or the surfaces with angle-dependent reflectances.}

\section{Conclusion}

In this paper, we have proposed Spectral MVIR, which is a novel method for joint 3D shape and spectral reflectance reconstruction. To jointly optimize each mesh vertex's position and spectral reflectance, we have built a cost optimization framework based on a multi-view and multispectral rendering model considering our image acquisition setup using a standard RGB camera and a low-cost LED bulb or LED projector. The light source positions also can be estimated jointly in the optimization, which makes our system based on Spectral MVIR free from geometric calibration. Experimental results using both synthetic and real-world data have demonstrated the potential of Spectral MVIR for the acquisition of a high-quality 3D model with accurate spectral reflectance property. We have also demonstrated that our Spectral MVIR can effectively be combined with the state-of-the-art Pro-Cam SSfM system and can produce a refined 3D shape with more accurate spectral reflectance.


\ifpeerreview \else
\fi

\bibliographystyle{IEEEtran}
\bibliography{egbib}

\ifpeerreview \else


\vspace{-4mm}
\begin{IEEEbiography}[{\includegraphics[width=1in,height=1.25in,clip,keepaspectratio]{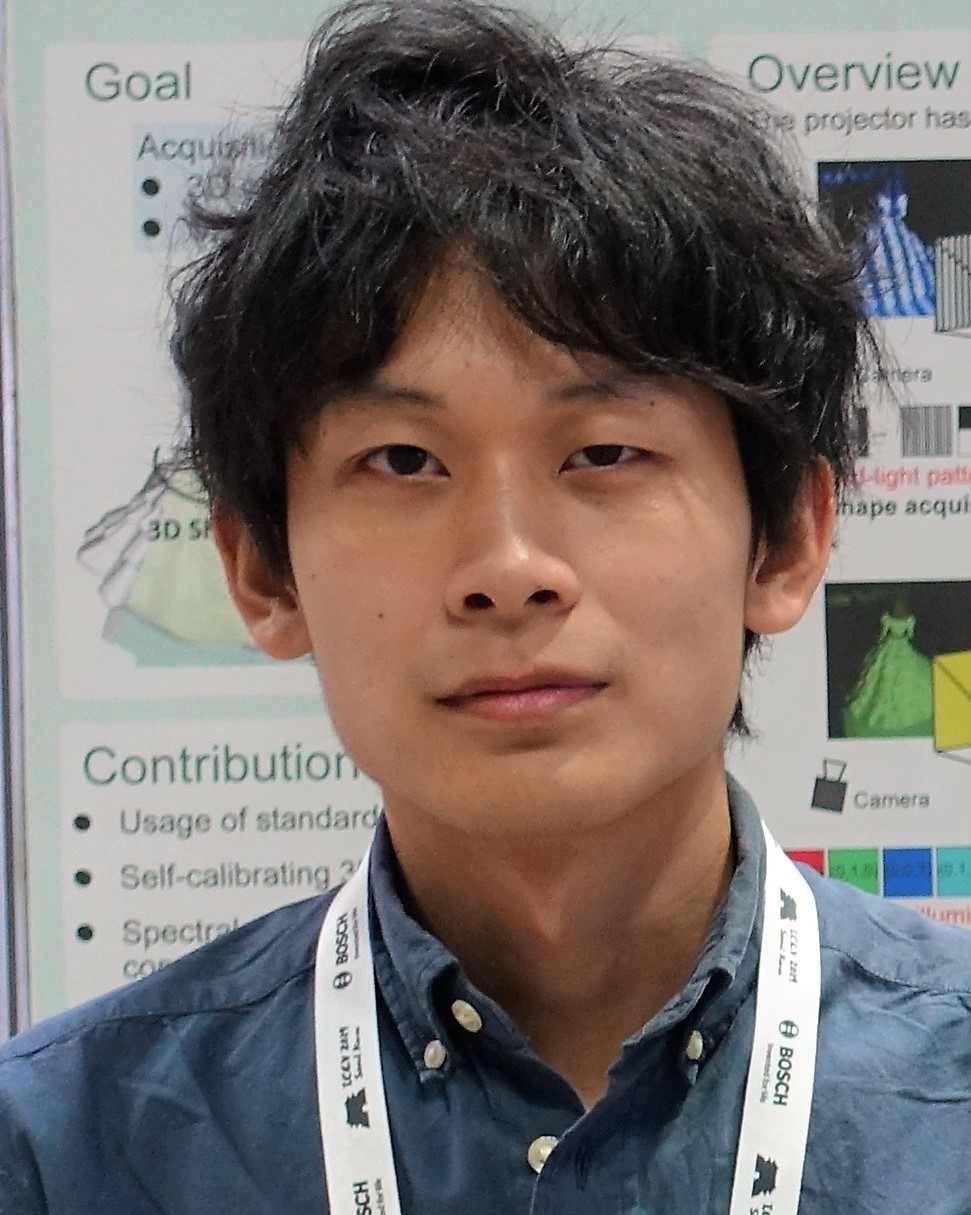}}]{Chunyu Li}
received the Bachelor of Engineering degree from the Software Institute, Nanjing University, in 2016, and the Master of Engineering degree from the Department of Systems and Control Engineering, Tokyo Institute of Technology, in 2019.
Currently he is a PhD candidate in the Department of Systems and Control Engineering, Tokyo Institute of Technology.
His research interests include 3D reconstruction (Structured-light, Structure from motion) and Hyperspectral Imaging.
\end{IEEEbiography}
\vspace{-4mm}
\begin{IEEEbiography}[{\includegraphics[width=1in,height=1.25in,clip,keepaspectratio]{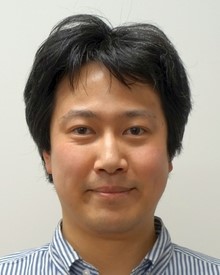}}]{Yusuke Monno}
received the B.E., M.E., and Ph.D degrees from Tokyo Institute of Technology, Tokyo, Japan, in 2010, 2011, and 2014, respectively. From Nov. 2013 to Mar. 2014, he joined the Image and Visual Representation Group at École Polytechnique Fédérale de Lausanne as a research internship student. He is currently an assistant professor with the Department of Systems and Control Engineering, School of Engineering, Tokyo Institute of Technology. His research interests are in both theoretical and practical aspects of image processing, computer vision, and biomedical engineering.
He is a member of IEEE.
\end{IEEEbiography}

\vspace{-4mm}
\begin{IEEEbiography}[{\includegraphics[width=1in,height=1.25in,clip,keepaspectratio]{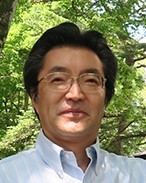}}]{Masatoshi Okutomi}
received the B.Eng. degree from the Department of Mathematical Engineering and Information Physics, the University of Tokyo, Tokyo, Japan, in 1981, and the M.Eng. degree from the Department of Control Engineering, Tokyo Institute of Technology, Tokyo, in 1983. He joined the Canon Research Center, Canon Inc., Tokyo, in 1983. From 1987 to 1990, he was a Visiting Research Scientist with the School of Computer Science, Carnegie Mellon University, Pittsburgh, PA, USA. He received the Dr.Eng. degree from Tokyo Institute of Technology, in 1993, for his research on stereo vision. Since 1994, he has been with Tokyo Institute of Technology, where he is currently a Professor with the Department of Systems and Control Engineering, the School of Engineering. 
\end{IEEEbiography}




\vfill

\end{document}